\documentclass[11pt]{article}
\usepackage[T1]{fontenc}
\usepackage[utf8]{inputenc}
\usepackage{lmodern}
\usepackage{microtype}
\usepackage[margin=1in]{geometry}
\usepackage{amsmath,amssymb,amsthm,mathtools}
\usepackage{booktabs,tabularx,array,multirow,longtable}
\usepackage{graphicx}
\usepackage{float}
\usepackage{xcolor}
\usepackage{xspace}
\usepackage{enumitem}
\usepackage{hyperref}
\usepackage{algorithm}
\usepackage{algpseudocode}
\usepackage{caption}
\usepackage{subcaption}
\graphicspath{{figs/}}

\definecolor{srblue}{RGB}{40,82,130}
\definecolor{srgreen}{RGB}{48,128,78}
\definecolor{srgray}{RGB}{95,101,110}
\hypersetup{colorlinks=true,linkcolor=srblue,citecolor=srgreen,urlcolor=srblue}
\setlist[itemize]{leftmargin=1.35em,itemsep=2pt,topsep=3pt}
\setlist[enumerate]{leftmargin=1.6em,itemsep=3pt,topsep=4pt}
\newcommand{\D}{\mathcal{D}}
\newcommand{\G}{\mathcal{G}}

\newcommand{\E}{\mathbb{E}}

\newcommand{\1}{\mathbf{1}}

\newcommand{\argmax}{\operatorname*{arg\,max}}

\newtheorem{proposition}{Proposition}
\newtheorem{definition}{Definition}

\newcommand{\longdksixplainT}{0.28\xspace}

\newcommand{\longdksixstructT}{0.92\xspace}
\newcommand{\longdksixD}{$+0.64$ $[+0.36,+0.90]$\xspace}

\newcommand{\longkctwoplainT}{0.00\xspace}

\newcommand{\longkctwostructT}{0.92\xspace}
\newcommand{\longkctwoD}{$+0.92$ $[+0.86,+0.98]$\xspace}

\newcommand{\longobsplainT}{0.00\xspace}

\newcommand{\longobsstructT}{0.62\xspace}
\newcommand{\longobsD}{$+0.62$ $[+0.36,+0.86]$\xspace}

\title{\textbf{StructRL: Model-Free Reinforcement Learning with Replay Graphs}}
\author{Ivo Nowak\\Hamburg University of Applied Sciences (HAW Hamburg)\\\texttt{ivo.nowak@haw-hamburg.de}}
\date{}

\begin{document}
\maketitle

\begin{abstract}
StructRL turns the experience collected during reinforcement learning into an explicit replay graph. Once the graph contains a path to the task goal, dynamic programming selects actions on the observed part of the state space; these actions form a graph policy that is defined only where the required transitions have been observed. StructRL transfers these actions to model-free learners: as action preferences for a DQN Q-model and as target directions along shortest observed paths for a SAC actor. For DQN on three sparse-reward MiniGrid tasks, final success rises from 0.10 to 0.90 (DoorKey), 0.10 to 0.80 (KeyCorridor) and 0 to 0.50 (ObstructedMaze); plain DQN does not catch up with three times the budget. For SAC on continuous PointMaze, graph-derived direction labels raise final success from 0.17 to 1.0 on seven matched runs that reached the goal. Distilling the graph policy into a fresh Q-model gives a DoorKey $6\times6$ controller that reaches the goal from 91\% of reachable states, against 38\% for plain DQN. Scaling experiments identify three bottlenecks: incomplete graph support, structural aliasing, and loss of graph information in high-dimensional inputs.
\end{abstract}


\section{Introduction}

Sparse-reward reinforcement learning often requires many interactions before a useful policy is found. In robotics and production control, the task goal and reward are often known even when the transition dynamics and the route to the goal are not \cite{plappert2018multigoal,gymnasiumrobotics2023,tang2025robotics,mayerhoff2026ppc}. StructRL uses the structure already contained in the experience collected during learning: observed transitions are connected into an empirical replay graph and evaluated relative to the task goal.

Consider the DoorKey task. An agent may already have observed how to pick up a key, open a door, and reach the goal from several states, while a standard replay buffer stores these observations only as individual training transitions. StructRL additionally connects them into a graph. Its nodes are \emph{structural states}: in DoorKey a structural state records the agent's position and direction, the object it carries and the state of the doors. This projection $\phi$ is specified from task knowledge and is not learned. Once the observed transitions form a path to the goal, dynamic programming can determine which observed action should be taken at each graph state on that path.

\begin{figure}[t]
\centering
\includegraphics[width=\linewidth]{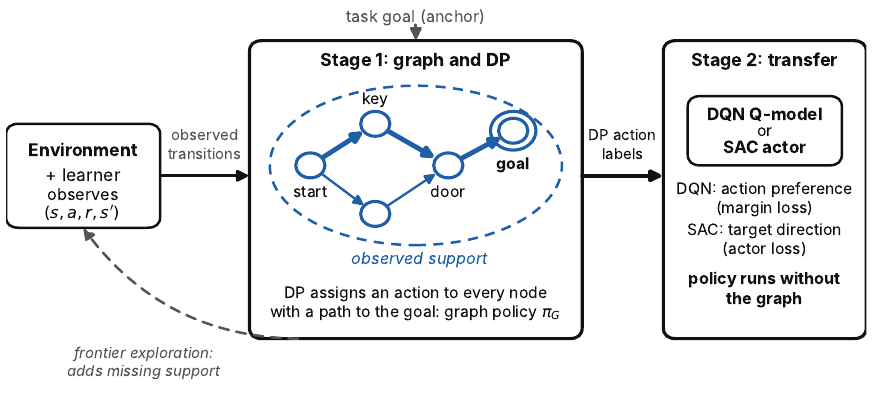}
\caption{StructRL in two stages. Stage 1: observed transitions form the replay graph, and dynamic programming gives the graph policy $\pi_G$ on the observed support (dashed region); outside it, no decision is made. Stage 2: the DP actions serve as labels that are transferred to the DQN Q-model or the SAC actor, which then acts without the graph. Frontier exploration (dashed arrow) is a side mechanism that adds missing observed transitions to the graph by steering the agent toward states with actions not yet observed.}
\label{fig:schema}
\end{figure}

This example illustrates the two stages of StructRL. First, the replay graph is generated from observed transitions and dynamic programming determines a restricted policy on the observed support. By \emph{restricted policy} we mean a policy that is defined only at structural states for which the graph contains a supported path to the goal; outside this support it makes no decision. We subsequently call this the \emph{graph policy}, $\pi_G$. Second, the graph-policy actions are used as DP action labels for a standard reinforcement-learning model. For DQN, they define action preferences for the Q-model; for SAC, they define target directions for the actor model. The graph is therefore an explicit solver on the part of the task that has been observed, while the Q-model or actor model provides a compact learned policy that can be executed without consulting the graph at every decision. Figure~\ref{fig:schema} summarizes these two stages; Table~\ref{tab:shared} lists which components are shared by DQN and SAC and which are specific to each.

The graph contains only observed transitions. Each observation $s$ is mapped to a structural state $z=\phi(s)$, and a graph edge is added only after the corresponding transition has been experienced. Planning can propagate goal information over these edges but never inserts an unobserved transition. StructRL does not use a supplied dynamics model or learned neural dynamics predictor, but its graph is an empirical transition model for planning. The learned policy is model-free: the DQN Q-model and the SAC actor are trained with their usual updates and act without the graph. The graph is an empirical transition model used during training to compute action labels and to guide support acquisition; \emph{model-free} in the title refers to the learned policy, not to an absence of any model in the full method. We call a task-relevant goal or event an \emph{anchor}; knowing an anchor specifies what counts as success, not how to reach it. When the goal is known, some DQN experiments also use a \emph{goal prior}: a Manhattan-distance hint that biases exploration toward the goal without providing a route, transition model, or shaped reward.

The experiments give three main results. First, ordinary StructRL improves DQN on all three primary MiniGrid tasks, and graph-derived direction labels improve the SAC actor on PointMaze after goal contact. Second, on states covered by the replay graph the graph policy is stronger than the learned Q-model, so the transfer of graph decisions into the learned model is incomplete. Third, the graph policy can be distilled into a fresh Q-model that reaches the goal from 0.91 of the reachable DoorKey $6\times6$ states, compared with 0.38 for plain DQN, when both are evaluated with the same random-action noise. These results do not establish generalization beyond observed support: randomized actions can return the agent to supported states, and the coverage evaluation measures success from reachable states rather than performance on unseen state distributions.

Controlled scaling experiments separate three limitations. Larger tasks can leave useful graph structure that the Q-model does not learn within the training budget. A structural projection that merges states requiring different actions destroys the restricted policy through aliasing. High-dimensional nuisance inputs can leave graph planning intact while preventing the Q-model from absorbing graph-derived labels; giving $\phi(s)$ to a separate Q-model branch removes this last failure. These failures separate three improvement targets for StructRL: support acquisition, construction of the structural representation, and transfer of the restricted policy to the learned model. Because each is measured separately, the results indicate which of these three components limits a graph-to-policy method, and not only whether it outperforms plain DQN.

The paper makes five contributions.
\begin{enumerate}[leftmargin=*]
\item We introduce StructRL, a two-stage policy-construction framework that first computes a restricted policy by dynamic programming on a generated replay graph and then transfers its action decisions to a standard RL model.
\item We instantiate this framework for two base models: action-preference transfer to a DQN Q-model and directional policy transfer to a SAC actor model, while their standard learning targets remain unchanged.
\item On the tested sparse-reward MiniGrid and PointMaze tasks, the resulting graph-informed policies improve over the corresponding plain DQN/SAC models when usable support is available. Matched controls separate graph-derived action information from goal information and, for the final DQN controller, from action randomization.
\item Scaling experiments identify incomplete support, structural aliasing and graph-to-Q-model feature access as separate failure modes; a two-branch Q-model repairs the feature-access failure.
\item A final-policy phase distils the graph policy: after training, a new Q-model is trained to imitate the graph actions on supported states and then acts without the graph. We evaluate state-space coverage and path length, including controls that separate graph-derived labels, distillation and randomization.
\end{enumerate}

\section{Related Work}

\paragraph{Replay-derived planning and graphs.}
Experience replay reuses past transitions \cite{lin1992}, while prioritized replay changes which transitions are sampled \cite{schaul2016}. Dyna and prioritized sweeping combine learning with explicit or learned transition models and predecessor-focused backups \cite{sutton1991,moore1993}; episodic and replay-derived methods likewise exploit stored trajectories or topology \cite{blundell2016mfec,lee2019ebu,eysenbach2019}. Topological Experience Replay (TER) orders replay by a backward graph search but does not derive a separate graph policy for interaction \cite{hong2022ter}. Search on the Replay Buffer, semi-parametric topological memory and sparse graphical memory plan over graphs built from stored observations for goal-conditioned navigation \cite{eysenbach2019,savinov2018,emmons2020}. ERLAM and Qgraph methods also connect replay structure with learned value or policy models \cite{zhu2020erlam,hoppe2020qgraph,hoppe2020industrial}. The distinction is therefore not the existence of replay graphs or graph search: both have substantial precedent. SoRB uses learned goal-conditioned distances to plan through replay observations and executes a local goal-conditioned controller; TER uses graph topology primarily to schedule replay backups. Qgraph and related work already connect graph information to value or policy learning. The particular configuration tested here derives action labels from observed one-step structural edges, uses graph guidance to acquire support, and transfers supported action rankings/directions to ordinary DQN/SAC networks without changing their Bellman targets. Table~\ref{tab:related} lists the closest of these methods. This combination is the experimental contribution; our results do not establish superiority to each cited method across its original settings.

\begin{table}[t]
\centering\footnotesize
\caption{Graph-based methods closest to StructRL: what the graph is built from and how it is used.}
\label{tab:related}
\begin{tabularx}{\textwidth}{@{}lXX@{}}
\toprule
Method & Graph built from & Use of the graph \\
\midrule
SoRB \cite{eysenbach2019} & replay observations, learned distances & waypoint planning for a goal-conditioned policy \\
TER \cite{hong2022ter} & replay transitions & order of replay backups \\
ERLAM \cite{zhu2020erlam} & episodic-memory states, observed transitions & value propagation; early guidance of a parametric model \\
Qgraph \cite{hoppe2020qgraph} & replay transitions, subgraph as simplified MDP & lower bounds that replace the TD target (QG-DDPG) \\
StructRL & observed edges between structural states $\phi(s)$ & action rankings (DQN) and directions (SAC) as auxiliary losses; Bellman targets unchanged \\
\bottomrule
\end{tabularx}
\end{table}

\paragraph{Action guidance and policy transfer.}
Demonstration-based methods such as DQfD, DDPGfD and AWAC transfer externally supplied action information into a learner \cite{hester2018dqfd,vecerik2017ddpgfd,nair2020awac}. StructRL may use a related ranking loss, but its labels are computed from the learner's own observed graph. The final-policy phase is also related to policy distillation, where a teacher policy is transferred into a student Q-model \cite{rusu2016,hinton2015}; here the teacher is the dynamic-programming graph policy.

\paragraph{Structural abstraction.}
A replay graph built on $z=\phi(s)$ trades fine state information for explicit connectivity. Exact state aggregation, lumpability and MDP homomorphisms characterize when such abstractions preserve the decision problem \cite{givan2003,ravindran2003,li2006}. StructRL treats task sufficiency of $\phi$ as an explicit assumption and tests failures caused by overly coarse aggregation rather than claiming that the representation is learned automatically.

\paragraph{Goals, exploration and curricula.}
Goal-conditioned RL often exposes the desired goal without revealing the route or transition dynamics \cite{plappert2018multigoal,gymnasiumrobotics2023}. Count-based exploration and graph frontiers address support acquisition from different signals \cite{bellemare2016pseudocount,yamauchi1997}, while Go-Explore and reverse-curriculum methods explicitly return to promising states or alter the start-state distribution \cite{ecoffet2021goexplore,florensa2017reverse,resnick2018backplay}. These mechanisms are complementary to StructRL: the task anchor says what matters, whereas replay determines which transitions are actually supported.

\section{Problem Formulation}
\label{sec:problem}

We consider an MDP
\begin{equation}
\mathcal M=(\mathcal S,\mathcal A,P,r,\gamma),
\end{equation}
with unknown transition kernel $P$. Interaction yields replay data
\begin{equation}
\D_t=\{(s_i,a_i,r_i,s'_i,d_i)\}_{i=1}^{N_t}.
\end{equation}
The DQN and SAC base learners use model-free updates, while StructRL explicitly constructs an \emph{empirical structural transition model} from observed data. Thus the full system combines model-free learning with model-based planning on recorded transitions. A structural projection
\begin{equation}
\phi:\mathcal S\rightarrow\mathcal Z,
\qquad z=\phi(s)
\label{eq:phi}
\end{equation}
groups observations into structural states for graph statistics.

A task interface additionally specifies an anchor criterion $\mathcal A_\star$. Our primary motivating case assumes that this criterion is available from the task or reward specification at $t=0$. A stricter fallback case registers the anchor only when the corresponding event is first observed. Crucially, knowing $\mathcal A_\star$ does not imply knowing a path to the anchor, a shortest-path distance through obstacles, the transition kernel, or an optimal action. The task tells StructRL which states are relevant; replay tells it which transitions have actually been observed. Optional geometric or topological priors constitute additional information and are evaluated separately rather than being folded into the definition of a known anchor.

The distinction between the environment MDP and the replay graph is essential. The replay graph is incomplete and policy-dependent. Missing edges are unsupported structural claims, not impossible transitions. In time-limited tasks, identical symbolic observations at different elapsed times can have different rewards or remaining horizons; the full Markov state must include that information for exact aggregation even if an implementation groups these observations together.

\paragraph{Assumption on $\phi$ and what graph planning guarantees.}
We treat the structural projection as an explicit assumption and separate three conditions: that $\phi$ is a valid abstraction, that the graph contains the relevant transitions, and that the planned actions can be executed.
\begin{definition}[Task-sufficient structural abstraction]\label{def:suff}
The projection $\phi$ is \emph{task-sufficient} if for all states $s_1,s_2$ with $\phi(s_1)=\phi(s_2)$ and all actions $a$: (i) $r(s_1,a)=r(s_2,a)$; (ii) $s_1\in\mathcal A_\star \Leftrightarrow s_2\in\mathcal A_\star$; and (iii) for every structural state $z'$, $\sum_{s':\phi(s')=z'}P(s'\mid s_1,a)=\sum_{s':\phi(s')=z'}P(s'\mid s_2,a)$.
\end{definition}
\begin{proposition}[Abstract MDP]\label{prop:abstract}
Assume bounded rewards and either $\gamma<1$ or a proper finite episodic task. If $\phi$ is task-sufficient, $\bar P(z'\mid z,a)=\sum_{s':\phi(s')=z'}P(s'\mid s,a)$ for any $s$ with $\phi(s)=z$ and the corresponding $\bar r(z,a)$ are well defined, and $\mathcal M_\phi=(\mathcal Z,\mathcal A,\bar P,\bar r,\gamma)$ is an MDP with $V^*(s)=\bar V^*(\phi(s))$ for all $s$. A policy $\bar\pi$ that is optimal in $\mathcal M_\phi$ yields the optimal policy $\pi(s)=\bar\pi(\phi(s))$ of the original MDP.
\end{proposition}
\emph{Proof.} See Appendix~\ref{app:proofs}.
This is the standard condition for exact state aggregation (model-irrelevance abstraction, lumpability, MDP homomorphism \cite{givan2003,ravindran2003,li2006}); we state it to fix what the graph guarantee below presupposes and make no claim of theoretical novelty. In a deterministic environment condition (iii) says that equal $\phi$ and equal action lead to the same next structural state. For the deterministic MiniGrid transition relation, the full symbolic key is intended to preserve structural successors. Exact task sufficiency for the reward objective, however, additionally requires the Markov state to include any relevant remaining time or step counter: MiniGrid rewards can depend on episode progress. The observed structural key omits elapsed time, so Proposition~\ref{prop:abstract} cannot be invoked unconditionally for the implemented reward objective. Coarser and continuous-state projections can also violate successor consistency, as in Section~\ref{sec:featureaccess} and on PointMaze. The replay graph $\G$ is the part of $\mathcal M_\phi$ that has been observed, and the graph guarantee is deliberately weak.
\begin{proposition}[Reachability on observed deterministic support]\label{prop:support}
Suppose the environment is deterministic, the structural projection preserves action-conditioned structural successors and anchor membership on states reachable from visited states, and the anchor is an absorbing goal for the planning objective. Every finite path of observed transitions from a supported structural state to the anchor is structurally executable from any represented state. Therefore the length of a shortest \emph{observed} path is an upper bound on the shortest feasible path length in the environment. A separate unit-cost shortest-path computation on the supported graph attains that observed path length, but the empirical reward-maximizing Bellman recursion in Eq.~\eqref{eq:graph-value} is not generally a shortest-path computation.
\end{proposition}
\emph{Proof.} See Appendix~\ref{app:proofs}.
\paragraph{Scope in the experiments.} On the deterministic MiniGrid layouts, Proposition~\ref{prop:support} gives executability of observed paths for the DQN labels (reward-based recursion, Eq.~\eqref{eq:graph-value}). The reward $1-0.9\,t/t_{\max}$ depends on the elapsed time $t$, which the key omits, so exact sufficiency of $\phi$ (Proposition~\ref{prop:abstract}) holds only up to this dependence. The SAC labels are unit-cost shortest observed paths, but the PointMaze cells are only partly transition-consistent ($C_\phi=0.59$), so the premises do not hold there and these labels are heuristics evaluated empirically. No result in Section~\ref{sec:results} is derived from the propositions.

The guarantee concerns supported-path feasibility and a separate unit-cost path objective, not a claim that the implemented reward-based graph policy always follows a shortest path: shorter routes that were never observed are invisible to the graph, which is why support acquisition is a separate component. For stochastic environments, the graph value is an estimate whose quality depends on the visit counts and we state no guarantee. A valid abstraction is not a complete support: Proposition~\ref{prop:abstract} concerns the former, Proposition~\ref{prop:support} what can be guaranteed from the part of the abstraction that has been observed. Three things can fail independently. The abstraction can fail, when states with different successors share a structural state; this is the case on PointMaze (Section~\ref{sec:hybrid}). The support can fail, when a needed transition has not been observed; unknown is not impossible. The execution can fail, when a planned structural step cannot be carried out reliably by the available actions; this is not evaluated here, since the discrete primitive actions of our benchmarks are treated as executable.

\section{StructRL}
\label{sec:method}

StructRL has two core stages: generation of a restricted policy from the empirical replay graph, followed by transfer of the resulting DP action decisions to the DQN Q-model or SAC actor model. Support acquisition extends the first stage when the graph does not yet contain a path to an anchor; the DQN and SAC mechanisms below instantiate the second stage differently.

\subsection{Empirical replay graph}
\label{sec:graph}

Let $z=\phi(s)$ denote a structural state. For $z,z'\in\mathcal Z$ and an action representation $u$, define
\begin{equation}
N_t(z,u,z')
=\sum_{i=1}^{N_t}
\1[\phi(s_i)=z,\;\psi(a_i)=u,\;\phi(s'_i)=z'].
\label{eq:counts}
\end{equation}
The replay graph is $\G_t=(V_t,E_t)$, where $V_t$ contains observed structural states and $(z,u,z')\in E_t$ only if the corresponding transition has been observed. Missing edges are unsupported, not impossible. For discrete actions, $\psi$ may be the identity; for continuous actions it may be a task-specific action representation or controller interface.

The graph is an empirical structural representation, not a supplied transition model. The neural learner can change the graph only by executing actions and adding new observed transitions. Value or policy estimates are never inserted into the graph as transition facts.

\paragraph{Graph value.}
Let $N_t(z,u)=\sum_{z'}N_t(z,u,z')$, and let $R_t(z,u,z')$ and $N^d_t(z,u,z')$ denote the accumulated reward and the number of terminations observed on $(z,u,z')$. A pair $(z,u)$ is \emph{supported} if $N_t(z,u)\ge n_{\min}$. On the supported pairs we solve the empirical Bellman equation by value iteration \cite{bertsekas1996},
\begin{align}
\hat Q(z,u) &= \hat r(z,u)+\gamma\sum_{z'}\hat P(z'\mid z,u)\bigl(1-\hat d(z,u,z')\bigr)\hat V(z'), \\
\hat V(z) &= \max_u \hat Q(z,u).
\label{eq:graph-value}
\end{align}
Here $\hat P(z'\mid z,u)=N_t(z,u,z')/N_t(z,u)$. The empirical reward is $\hat r(z,u)=\sum_{z'}R_t(z,u,z')/N_t(z,u)$, and $\hat d(z,u,z')=N^d_t(z,u,z')/N_t(z,u,z')$ is the observed termination rate. An \emph{anchor transition} is a supported pair $(z,u)$ with an observed positive reward, and $\mathcal Z_A$ denotes the \emph{anchor-reachable} structural states, i.e.\ those from which an anchor transition can be reached over supported pairs. The graph policy is $\pi_G(z)=\argmax_{u\text{ supported}}\hat Q(z,u)$ for $z\in\mathcal Z_A$. Rewards are nonnegative in all our environments, so supported actions outside $\mathcal Z_A$ have value zero; they carry no goal information, $\pi_G$ is not defined there, and no label is derived from them. We call a state \emph{supported} if it has at least one supported action; $\mathcal Z_A$ is the set of supported states with a supported path to an anchor transition. The graph controller that is \emph{executed} (in guided DQN episodes and in the final-policy controllers) is $\pi_{\mathrm{exec}}(z)=\pi_G(z)$ for $z\in\mathcal Z_A$, the first supported action for a supported state outside $\mathcal Z_A$, and undefined at states without a supported action. This maximizes the empirical discounted reward criterion, which may differ from a minimum-step path objective; in particular, path-length terminology elsewhere refers to a separate diagnostic or a unit-cost interpretation rather than an unconditional guarantee for $\pi_G$.

\paragraph{Frontier value.}
To direct the agent toward unexplored parts of the graph, we propagate the novelty reward $r_\nu(z,u)=\beta/\sqrt{N_t(z,u)+1}$ with a max-plus recursion,
\begin{equation}
Q_\nu(z,u)=\max\Bigl\{r_\nu(z,u),\;\gamma_\nu\sum_{z'}\hat P(z'\mid z,u)\,V_\nu(z')\Bigr\},\qquad V_\nu(z)=\max_u Q_\nu(z,u),
\label{eq:frontier-value}
\end{equation}
and assign $Q_\nu(z,u)=\beta$ to actions not yet tried in $z$. The frontier policy is $\pi_\nu(z)=\argmax_u Q_\nu(z,u)$. A pair visited since the last planning step is capped at $\beta/\sqrt{N_t(z,u)+1}$, which avoids repeating a stale action within an episode.

\subsection{Anchor-relative planning and support acquisition}

StructRL separates graph construction, anchor-relative planning, and exploration.

\paragraph{Anchor.}
An anchor is a task-relevant state or event. In the primary practical setting it is supplied by a reward or goal specification before learning starts. In the fallback setting it is registered only when the event is first observed. Anchor knowledge alone does not determine an action before an empirically supported path exists. The original DQN and SAC performance blocks register the anchor online at first reward contact. The known-anchor DQN block exposes the goal position at $t=0$ and separately adds a Manhattan-distance prior for frontier weighting; this prior is additional information beyond merely naming the anchor.

\paragraph{Anchor-relative planning.}
Once observed support connects a structural state to an anchor transition, the empirical Bellman solution in Eq.~\eqref{eq:graph-value} induces a supported graph action $\pi_G(z)$. This is the dynamic-programming component of StructRL: information associated with the anchor is propagated over already observed predecessor structure. No new graph edge is created by planning. We call $\pi_G$ the restricted policy because it is defined only on the anchor-reachable states $\mathcal Z_A$; outside this observed support it makes no decision. All labels transferred to the learner (Sections~\ref{sec:dqn-margin}, \ref{sec:final-policy} and \ref{sec:sac}) are restricted to $\mathcal Z_A$.

\paragraph{Support acquisition.}
If no supported anchor path is available, StructRL can use the frontier value in Eq.~\eqref{eq:frontier-value} to navigate through known graph structure toward under-sampled state-action pairs and test them in the environment. The resulting transition, not the planner, creates the new graph edge. This makes the exploration graph-based without assuming that the graph can plan through unobserved dynamics. Optional task priors can bias frontier priorities when anchor information is available. We treat these as a separate information layer: a simple coordinate-distance prior uses only a task-defined notion of proximity, whereas a wall-map or shortest-path prior additionally assumes environment topology. Neither is required for the empirical replay graph itself.

\begin{algorithm}[t]
\caption{StructRL}\label{alg:structrl}
\begin{algorithmic}[1]
\State Initialize learner, replay memory $\D$, empty graph $\G$
\State Initialize anchor set $\mathcal A$ from task information if supplied, otherwise $\mathcal A\leftarrow\varnothing$
\For{each environment step $t$}
  \If{the learner-specific planning schedule triggers}
    \State Recompute supported graph values $\hat Q$ and frontier values $Q_\nu$ on $\G$
  \EndIf
  \State $g_t\leftarrow\pi_{\mathrm{exec}}(\phi(s_t))$ once an anchor transition has been observed, else $\varnothing$ \Comment{$\pi_G$ on $\mathcal Z_A$, first supported action elsewhere}
  \If{graph control is active and $g_t\neq\varnothing$ and graph guidance is executed}
    \State $a_t\leftarrow g_t$ \Comment{DQN graph action}
  \ElsIf{graph control is active and $g_t=\varnothing$}
    \State $a_t\leftarrow\pi_\nu(\phi(s_t))$ \Comment{support acquisition / frontier action}
  \Else
    \State $a_t\sim$ learner policy
  \EndIf
  \State Execute $a_t$, observe $(r_t,s_{t+1},d_t)$
  \State $\D\leftarrow\D\cup\{(s_t,a_t,r_t,s_{t+1},d_t)\}$
  \State Add $(\phi(s_t),\psi(a_t),r_t,\phi(s_{t+1}),d_t)$ to $\G$
  \If{the anchor criterion is observed online}
    \State Add the corresponding anchor to $\mathcal A$
  \EndIf
  \State Update the base model from $\D$ \Comment{learner-specific graph-to-action term may be added}
\EndFor
\end{algorithmic}
\end{algorithm}

Algorithm~\ref{alg:structrl} states the common logic. The graph is updated after every environment step, while planning follows a model-specific schedule (Appendix~\ref{app:impl}). In the primary known-anchor setting the anchor is registered at $t=0$, but the graph cannot infer a route until interaction has produced supported edges. In the fallback setting the anchor is registered only after reward contact. In both cases, support acquisition and anchor-relative planning use only observed graph edges.

\subsection{Structural representation and temporal granularity}
The replay graph is built on structural states $z=\phi(s)$. If $\phi$ is too coarse, states that require different decisions are merged; if it is too fine, useful support is fragmented. Continuous actions may also require a graph decision to persist for several environment steps. These choices affect implementation but not the two-stage definition of StructRL; the tested choices are described in Appendix~\ref{app:impl}.

\begin{table}[t]
\centering\small
\caption{Learner-specific components. Both learners build the graph from observed transitions, restrict labels to $\mathcal Z_A$ and leave the Bellman targets unchanged.}
\label{tab:shared}
\begin{tabularx}{\textwidth}{@{}lXX@{}}
\toprule
 & DQN (discrete actions) & SAC (continuous actions) \\
\midrule
Graph value & reward-based recursion, Eq.~\eqref{eq:graph-value} & unit-cost shortest observed path \\
Label & set of graph-optimal actions at $z\in\mathcal Z_A$ & direction to the next cell on the path, goal position in the last cell \\
Loss on the learner & large-margin ranking loss next to the TD loss & squared direction loss on the actor mean \\
\bottomrule
\end{tabularx}
\end{table}

\subsection{DQN as base model}
\label{sec:dqn}
For discrete actions we use Double DQN \cite{mnih2015dqn}. Before the graph contains a supported path to the goal, graph-guided episodes use frontier actions to acquire missing support. Once a path exists, graph-guided episodes can follow the graph policy for part of the episode; otherwise the DQN Q-model acts normally. The graph changes behavior only through observed actions and never changes the Double-DQN target. Schedules, random-action rates and other implementation settings are given in Appendix~\ref{app:impl} and Table~\ref{tab:hyper-dqn}.

\subsection{Graph-derived action-preference transfer for DQN}
\label{sec:dqn-margin}
Direct graph actions improve behavior only while graph guidance is active. To transfer the graph preference into the Q-model, we additionally evaluate a global margin objective on structural states with a supported path to the anchor. Let $T(z)$ be the set of supported graph-optimal actions at node $z$. For replay observations with $\phi(s)=z$, the auxiliary loss is
\begin{equation}
L_{\mathrm{margin}}=
\lambda\,\max\!\left(0,
\max_{a\notin T(z)}[Q_\theta(s,a)+m]
-
\max_{a\in T(z)}Q_\theta(s,a)
\right),
\label{eq:dqn-margin}
\end{equation}
The ordinary Double-DQN update is unchanged; the margin transfers an action ranking, not a calibrated graph value. Parameter values and the matched shuffled-label control are specified in Appendix~\ref{app:impl}.

\subsection{Final policy phase: graph-policy distillation}
\label{sec:final-policy}
The Q-model trained during environment interaction is referred to as the \emph{online Q-model}. The final-policy experiment asks whether the graph policy can be transferred into a new model after interaction has ended. We freeze the replay graph and replay buffer, compute the graph-policy actions once more, and train a freshly initialized Q-model on these actions as hard classification targets. No new reward or environment interaction is used. At evaluation, the resulting Q-model can be combined with a fixed share of uniformly random actions; the same randomization is applied to plain and online Q-model controls so that the effect of random actions is not attributed to distillation. Soft targets and support-aware randomization are additional controls. The distillation loss, optimization settings and exact randomization definitions are given in Appendix~\ref{app:final-details}.

\subsection{SAC as base learner}
\label{sec:sac}
For continuous actions we use SAC \cite{haarnoja2018} with two critics, automatic entropy tuning and the target entropy $-\dim\mathcal A$. A single continuous action rarely moves the agent to a new structural node, so the two stages use different mechanisms: before the anchor, graph decisions are executed as persistent prototype actions; after the anchor, the graph acts only through an auxiliary loss on the actor. Unlike in the DQN setting, the SAC policy never executes $\pi_G$ directly.

\paragraph{Structural state and action representation.}
The structural state $\phi(s)$ is the cell of a regular grid with side length $0.5$ over the agent position $p(s)$. Actions are represented by $K=9$ prototypes $u_0,\dots,u_8$: the zero action and the eight compass directions, scaled so that $\|u_k\|_\infty=1$. The representation $\psi(a)$ is the index of the nearest prototype.

\paragraph{Stage 1: frontier policy.}
On the graph over $(\phi,\psi)$ the frontier values are
\begin{equation}
Q_\nu(z,u)=\max\Bigl(\frac{\beta}{\sqrt{N(z,u)+1}},\;\gamma_\nu\sum_{z'}\hat P(z'\mid z,u)\,V_\nu(z')\Bigr),\qquad V_\nu(z)=\max_u Q_\nu(z,u),
\label{eq:sac-frontier}
\end{equation}
where $N(z,u)$ is the visit count, $\hat P$ the empirical transition frequency, and pairs that were never tried receive the optimistic value $\beta$. Values are recomputed every 500 steps and at the end of every episode. Actions are taken in blocks of $H=10$ steps. At the start of a block, with probability $p_{\mathrm{explore}}$, the frontier policy selects $u^\star=\argmax_u Q_\nu(\phi(s),u)$, which is executed for the whole block with clipped Gaussian noise drawn at every step; otherwise the SAC policy acts for the block. Stage~1 ends at the anchor, the first transition with reward $r>0.5$.

\paragraph{Stage 2: graph-direction loss.}
The SAC labels come from a unit-cost replay graph over the cells of $\phi$ (an edge is kept from its first observation); a backward breadth-first search from the goal edge gives shortest observed paths. After the anchor, the graph contains an observed path to the anchor from a growing set $\mathcal S_G=\mathcal Z_A$ of cells. For $s$ with $\phi(s)\in\mathcal S_G$, let $z^+(s)$ be the next node on a shortest observed path to an anchor transition, and let $c(z)$ be the center of cell $z$. The graph direction is
\begin{equation}
a_G(s)=\frac{y(s)-p(s)}{\|y(s)-p(s)\|_\infty},\qquad
y(s)=\begin{cases}c\bigl(z^+(s)\bigr) & \text{if } \phi(s) \text{ is not the last node of the path},\\ g(s) & \text{otherwise,}\end{cases}
\label{eq:sac-label}
\end{equation}
where $g(s)$ is the goal position contained in the observation. Let $\mu_\theta(s)$ be the mean of the Gaussian policy before the $\tanh$ squashing. For a minibatch $\mathcal B$ with supported part $\mathcal B_G=\{s\in\mathcal B:\phi(s)\in\mathcal S_G\}$, the actor minimizes
\begin{equation}
L_\pi(\theta)=\E_{s\in\mathcal B,\;\tilde a\sim\pi_\theta(\cdot\mid s)}\Bigl[\alpha\log\pi_\theta(\tilde a\mid s)-\min_{i=1,2}Q_{\omega_i}(s,\tilde a)\Bigr]
+\lambda\,\frac{1}{|\mathcal B_G|}\sum_{s\in\mathcal B_G}\bigl\|\tanh\mu_\theta(s)-a_G(s)\bigr\|_2^2 .
\label{eq:sac-loss}
\end{equation}
The second term is set to zero when $\mathcal B_G=\varnothing$. Before the anchor, $\mathcal B_G$ is empty and \eqref{eq:sac-loss} is the standard SAC actor loss. The critic loss, the soft Bellman target and the entropy update are those of SAC and are not affected by the graph. The hyperparameters are listed in Table~\ref{tab:hyper-sac}.

\paragraph{SAC configurations.}
The SAC baseline (\texttt{P}) uses no graph. Frontier-only SAC (\texttt{F}) uses graph-based support acquisition before the anchor but no graph-direction loss afterwards. StructRL-SAC (\texttt{FL}) uses both stages. The random-label control (\texttt{FLR}) is identical to StructRL-SAC except that $a_G(s)$ is replaced by a fixed random direction per node, $(\cos\vartheta_z,\sin\vartheta_z)/\|(\cos\vartheta_z,\sin\vartheta_z)\|_\infty$ with $\vartheta_z$ uniform on $[0,2\pi)$. StructRL-SAC and the random-label control therefore have the same support and identical pre-anchor behavior; they differ only in the post-anchor direction label.

\subsection{The role of $\phi$ and feature access}
\label{sec:phi-role}
The structural projection $\phi$ has two distinct roles. First, it defines the nodes of the replay graph, $z=\phi(s)$, and therefore decides which observed states are treated as identical for planning. Second, in the original DQN Q-model the raw observation $o$ is the only input; $\phi(s)$ does not enter the Q-model, and the graph acts on the learner only through labels, guided episodes and the margin loss. With few irrelevant observation dimensions this suffices. With many irrelevant dimensions the joint input layer must find the structurally relevant coordinates among the nuisance, and graph-derived labels do not make this selection easier.

We therefore evaluate a \emph{two-branch} Q-model that receives the raw observation and $\phi(s)$ through separate input branches before combining them in the DQN trunk. The graph, planning and graph-action transfer are unchanged. This tests whether the failure under nuisance input is caused by poor access to the structural features rather than by the replay graph itself. Controls that isolate the branches, detach gradients or add an explicit DP head, and a gradient-cosine diagnostic, are described in Appendix~\ref{app:phi2}.

\paragraph{Transition consistency of $\phi$.}
Task sufficiency (Definition~\ref{def:suff}) can be partly checked on data. For a structural state-action pair $(z,u)$, let $\hat z^{+}(z,u)$ be its most frequent observed successor and
\begin{equation}
c(z,u)=\frac{N\bigl(z,u,\hat z^{+}(z,u)\bigr)}{N(z,u)}
\label{eq:cphi}
\end{equation}
the share of its transitions that end there. The \emph{transition consistency} $C_\phi$ is the average of $c(z,u)$ over the counted pairs, i.e. the share of structural decisions whose realized successor is the most frequent one targeted by planning. $C_\phi=1$ means that every structural decision has a single observed outcome, as condition (iii) of Definition~\ref{def:suff} requires in a deterministic environment; there, $C_\phi<1$ is direct evidence that $\phi$ does not preserve the observed transition relation, whereas $C_\phi=1$ on observed data does not prove the condition for transitions not yet observed, and in a stochastic environment $C_\phi<1$ can also result from the randomness of the environment itself. A value well below one means that the successor targeted by planning is only one of several outcomes of the same structural decision, so that a graph path is not reliably executable. In DQN, $C_\phi$ is the share of the steps chosen by the graph policy during evaluation whose realized successor equals $\hat z^{+}$. On PointMaze it is measured without training, on random compass actions held for ten steps; only transitions that change the cell are counted, and $c(z,u)$ is averaged over pairs with at least two such transitions.

\subsection{Determining $\phi$}
\label{sec:phi-determine}
The main experiments use a $\phi$ that follows from the task: for the discrete grid tasks the agent position and direction, the carried object and the grid contents, for PointMaze a regular grid over the position. Such a $\phi$ is task-sufficient (Definition~\ref{def:suff}) on the deterministic MiniGrid tasks, and Section~\ref{sec:hybrid} shows that it is only partly transition-consistent on PointMaze. In general $\phi$ is determined by a short loop: candidate variables are chosen from the task (those that decide which action is possible and useful, not timestamps, sensor noise or history), continuous variables are discretized, the interaction or a simulator produces transitions, and the consistency of reward and successor within each structural state-action pair $(z,a)$ is measured; high inconsistency calls for a finer $\phi$, and very low inconsistency allows removing variables. The loop is a data-based approximation of the task sufficiency of Definition~\ref{def:suff}. Its first step is used throughout; the selection and refinement steps are evaluated in controlled settings.

\paragraph{Specification and test.} $\phi$ is specified from the task and then tested on data with the transition consistency $C_\phi$ \eqref{eq:cphi}: $C_\phi<1$ in a deterministic environment shows that $\phi$ merges states with different successors. Coarsened projections (without heading, or position and object only) fail this test and destroy the graph policy (Section~\ref{sec:scaling}).

\paragraph{Automatic selection and refinement.} We also test whether nuisance dimensions can be removed automatically and whether a detected inconsistency can trigger re-aggregation under a finer predefined projection. With 1000 synthetic nuisance dimensions the selection rule reaches stable success in 5 of 5 seeds with median AUC 0.72, against 0.92 for the hand-built $\phi$. These tests show feasibility but do not solve automatic representation learning: the selection rule still requires a structural key, and the refinement experiment does not improve over the hand-built representation. The rules, the corrections made after two failed versions and the results are reported in Appendix~\ref{app:phi2}.

Constructing $\phi$ therefore shifts part of the problem from policy learning to representation design. We do not evaluate this cost in realistic applications with many sensors, hidden process states and delays; learning $\phi$ directly from replay or simulator data remains future work.

\section{Experimental Design}
\label{sec:experiments}

\subsection{Evaluation design and decision rules}
The main configurations (Table~\ref{tab:variants}) are named by the information added to the plain model:
\begin{table}[H]
\caption{Main configurations used in the text. Internal implementation codes are retained only for reproducibility.}
\label{tab:variants}
\centering\small
\begin{tabular}{lll}
\toprule
Name in text & Added information & Internal code\\
\midrule
DQN baseline & none & \texttt{plain}\\
StructRL & replay graph and graph guidance & \texttt{struct}\\
StructRL+Prior & StructRL plus goal-distance prior & \texttt{struct\_ka}\\
StructRL+Transfer & StructRL+Prior plus graph-action transfer & \texttt{struct\_mgkahg}\\
SAC baseline & none & $P$\\
StructRL-SAC & frontier support acquisition plus graph direction labels & $FL$\\
\bottomrule
\end{tabular}
\end{table}
The experiments use matched or paired controls and keep task families separate. Within each comparison, arms use matched environment-step budgets (typically 60k--150k steps in the primary task blocks), selected in part for computational feasibility and not to demonstrate baseline convergence; runtime can exceed 30 minutes for the scaling runs of Section~\ref{sec:scaling}, and some coverage evaluations were resumed across queue windows (Appendix~\ref{app:protocol}). All runs used CPU only (two cores of an Intel Xeon at 2.10 GHz, 7 GB memory, one thread per run; Python 3.13, PyTorch 2.14, Stable-Baselines3 2.9, MiniGrid 3.2, Gymnasium 1.4, Gymnasium-Robotics 1.4). The main DQN comparison tests the DQN baseline against StructRL. The known-goal experiments compare StructRL, StructRL+Prior, and StructRL+Transfer. Held-out DQN tasks test the same settings without retuning. The SAC experiments compare the SAC baseline, frontier-only SAC, StructRL-SAC, and matched controls with the same support but without graph-derived direction labels. Scaling experiments vary task size, the structural projection $\phi$, and nuisance observation dimensions separately. For each experiment block, the arms, seeds and decision rule were written down before the runs; they are listed in Appendix~\ref{app:protocol} together with the deviations from this protocol.

Feature-access experiments compare a joint Q-model with a two-branch Q-model that receives $\phi(s)$ directly. Additional controls remove graph episodes, detach the raw branch, or add an explicit graph-action head. TER is evaluated with the official graph, search and encoder components inside the same DQN implementation. Final-policy experiments compare the online Q-model, distilled Q-models, graph policies and matched randomization controls over reachable DoorKey states.

\subsection{TER and graph-policy comparisons}
\label{sec:q7}
The TER block compares uniform replay, TER, ordinary StructRL and graph-adapted StructRL on DoorKey $6\times6$, UnlockPickup and DoorKey $8\times8$. Separate hybrid evaluations combine the executed graph controller $\pi_{\mathrm{exec}}$ on supported states with either the Q-model or a uniformly random fallback elsewhere. The primary off-support contrast is the paired success difference between those two fallbacks; the full arm definitions, seeds and decision rules are in Appendix~\ref{app:protocol}.

\subsection{Final-policy coverage}
\label{sec:q8}
Let $\mathcal S_{\mathrm{reach}}$ be the states reachable from the initial state in deterministic DoorKey, $\mathcal S_{\mathrm{sup}}$ the supported states and $\mathcal Z_A\subseteq\mathcal S_{\mathrm{sup}}$ the anchor-reachable states. We measure the shares $C_G$ of reachable states that are supported and $C_A$ of those in $\mathcal Z_A$, and the policy coverage $C_\theta$, the share of reachable states from which a controller reaches the goal within the step limit. DoorKey $6\times6$ is evaluated exhaustively; DoorKey $8\times8$ uses the same fixed sample of 600 reachable states for all controllers within a seed. The evaluation compares the Q-model trained during interaction, distilled Q-models, random-action controls and $\pi_{\mathrm{exec}}$ with random fallback. Wherever the graph is executed on its own (graph-only rollouts, hybrid controllers, random fallback) this is $\pi_{\mathrm{exec}}$ of Section~\ref{sec:graph}; labels and distillation targets are derived from $\pi_G$. The decision rules (thresholds for distillation, randomization and the support-aware variant, and the rules E1 and E2 of the randomization control) are given in Appendix~\ref{app:protocol}.

\subsection{Environments}
Table~\ref{tab:envs} lists the environments. MiniGrid tasks \cite{chevalierboisvert2023minigrid} have seven discrete actions and sparse terminal reward. PointMaze \cite{gymnasiumrobotics2023} is a continuous-control navigation task with a two-dimensional force and sparse goal reward.

\begin{table}[t]
\centering\small
\caption{\textbf{Environments.} Observation and action details are given in Appendix~\ref{app:envs}.}
\label{tab:envs}
\begin{tabularx}{\linewidth}{@{}>{\raggedright\arraybackslash}p{2.55cm}>{\raggedright\arraybackslash}p{2.3cm}>{\raggedright\arraybackslash}X>{\raggedright\arraybackslash}p{3.3cm}@{}}
\toprule
Environment & Actions & Task & Purpose \\
\midrule
DoorKey 6x6 & discrete (7) & pick up the key, open the door, reach the goal & sparse sequential discovery \\
KeyCorridor S3R2 & discrete (7) & find the key, open a locked door, pick up the object behind it & longer sparse sequence \\
ObstructedMaze 1Dl & discrete (7) & open a locked door with the visible key and pick up the object behind it & hard sparse discovery \\
UnlockPickup, KeyCorridor S3R3 & discrete (7) & unlock a door and pick up an object; larger corridor with three rooms & held-out tasks \\
PointMaze (M, Large) & continuous, 2-D force & move a ball through a maze to a goal cell & continuous sparse navigation \\
SearchGrid & discrete (5) & visit $m$ ordered subgoal cells in an $n\times n$ grid with walls & controlled discovery family (planner only, no learner) \\
\bottomrule
\end{tabularx}
\end{table}

The experiments use small symbolic MiniGrid tasks and low-dimensional PointMaze states, with main training budgets of 50k--150k environment steps. They therefore test the mechanism rather than image-based or large-scale scalability. The published evaluations of TER and SoRB use image observations, and TER is trained for millions of environment steps; we therefore do not compare scores with those papers.

\subsection{DQN Q-model setup}
\label{sec:dqn-setup}
The primary DQN experiments use DoorKey $6\times6$, KeyCorridor S3R2 and ObstructedMaze 1Dl with ten seeds per task and budgets of 60k, 150k and 100k environment steps. All configurations use the same Double+Dueling+$n$-step DQN Q-model and the same evaluation schedule; only the graph components differ. The DQN baseline uses no graph, ordinary StructRL uses graph-based discovery and graph guidance, StructRL+Prior adds the goal prior defined above, and StructRL+Transfer additionally transfers graph-derived action preferences into the Q-model. Exact hyperparameters, internal code names and the prior weighting are given in Appendix~\ref{app:impl} and Table~\ref{tab:hyper-dqn}.

\paragraph{Held-out tasks.}
UnlockPickup and KeyCorridor S3R3 use seeds 700--709 for 100k steps without retuning. The same known-anchor arms are applied without changing the prior or margin settings.

\subsection{SAC actor-model setup}
\label{sec:sac-setup}
The SAC experiments use PointMaze M and Large. The main comparison tests plain SAC against StructRL-SAC, while matched controls keep pre-goal exploration and graph support fixed but remove or randomize the graph-derived direction labels. The known-goal PointMaze Large block uses ten seeds; post-goal label contrasts are evaluated only on matched runs that actually reach the goal, because no graph-direction label exists before goal contact. Exact internal arm names, seed sets, evaluation schedules and hyperparameters are given in Appendix~\ref{app:protocol} and Table~\ref{tab:hyper-sac}.

\section{Results}
\label{sec:results}

\begin{table}[H]
\caption{Main end-to-end comparisons. DQN rows report the primary training evaluations. PointMaze reports the SAC improvement over plain SAC and the matched post-goal-contact label comparison. Reachable-state coverage is a separate final-policy evaluation.}
\label{tab:main-comparison}
\centering\small
\begin{tabular}{llrr}
\toprule
Task / evaluation & Metric & Baseline & StructRL result\\
\midrule
DoorKey $6\times6$ & final success & 0.10 & 0.90\\
DoorKey $6\times6$ & AUC & 0.01 & 0.41\\
KeyCorridor S3R2 & final success & 0.10 & 0.80\\
KeyCorridor S3R2 & AUC & 0.04 & 0.44\\
ObstructedMaze & final success & 0.00 & 0.50\\
ObstructedMaze & AUC & 0.00 & 0.23\\
\midrule
PointMaze Large & evaluation AUC & plain SAC & $+0.28$ $[+0.15,+0.41]$\\
PointMaze Large, goal-contact runs & final success & no labels: 0.17 & graph labels: 1.00\\
\midrule
DoorKey $6\times6$, reachable states & coverage & plain DQN: 0.38 & final policy: 0.91$^\dagger$\\
\bottomrule
\multicolumn{4}{l}{\footnotesize $^\dagger$StructRL+Transfer final controller; ordinary StructRL reaches 0.90.}\\
\end{tabular}
\end{table}

Table~\ref{tab:main-comparison} summarizes the main end-to-end evidence before the mechanism controls. The DQN rows measure learning from the nominal task start. The PointMaze rows show that graph-derived action information also improves SAC, with the post-anchor comparison conditioned on runs that reach the goal. The coverage row asks a different question: how broadly the final DoorKey controller succeeds across reachable states. The following subsections answer three questions: does StructRL work, why does it work, and where does it fail? Each paragraph begins with its question; full results with intervals and sample sizes are in Appendix~\ref{app:details}.

\subsection{Does graph guidance improve learning?}
\begin{figure}[t]
\centering
\includegraphics[width=0.88\linewidth]{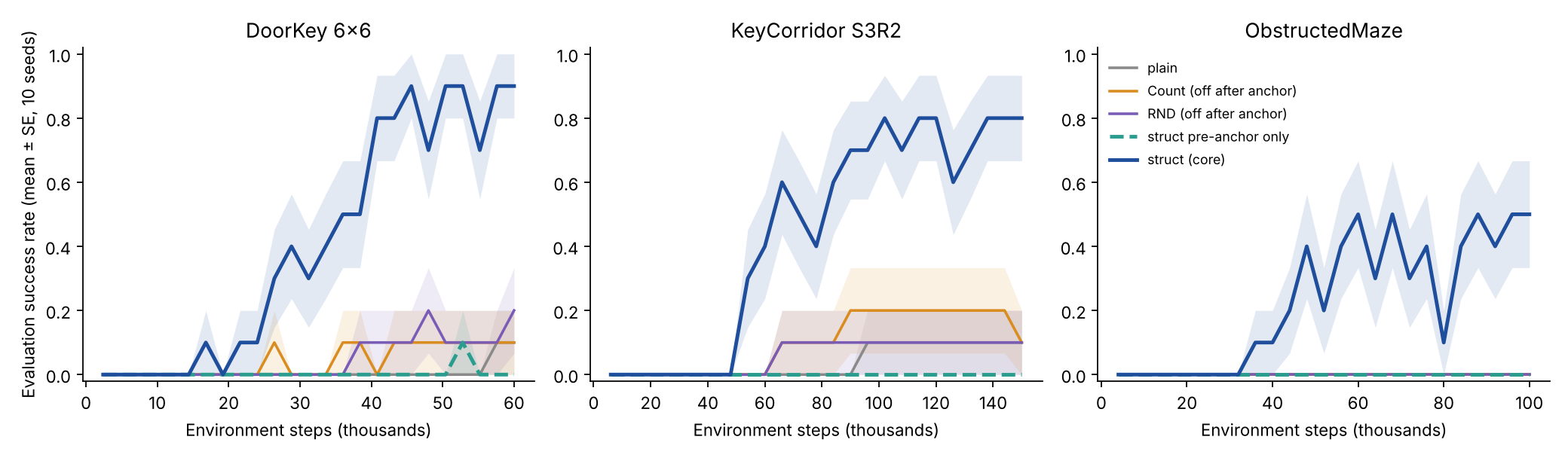}
\caption{Raw, unsmoothed DQN evaluation success over environment steps on three tasks (seeds 500--509, mean $\pm$ SE; one deterministic evaluation episode per checkpoint).}
\label{fig:dqn-curves}
\end{figure}

\paragraph{DQN on three sparse-reward tasks.} Does explicit graph structure improve DQN? StructRL reaches the anchor, i.e.\ the first observed goal transition, in 10/10 seeds on all three tasks, against 6/10, 4/10 and 1/10 for plain DQN, and its mean evaluation AUC is 0.41, 0.44 and 0.23 against 0.01, 0.04 and 0.00 (Table~\ref{tab:main-comparison}, Figure~\ref{fig:dqn-curves}). Discovery-only StructRL also reaches the anchor in 10/10 seeds but has AUC 0.00: reaching the anchor is not sufficient, the graph actions must reach the learner. Details and paired intervals are in Appendix~\ref{app:details}.

\paragraph{Three times the training budget.} Does the baseline catch up with more interaction? Both arms of the three DQN blocks were rerun with three times the environment steps on the same seeds (Table~\ref{tab:long}). The tail success of plain DQN rises from 0.04 to \longdksixplainT{} on DoorKey $6\times6$ and stays at \longkctwoplainT{} and \longobsplainT{} on KeyCorridor S3R2 and ObstructedMaze 1Dl, against \longdksixstructT{}, \longkctwostructT{} and \longobsstructT{} for StructRL; the paired tail differences are \longdksixD, \longkctwoD{} and \longobsD. The performance advantage persists over the extended budget, and the baseline shows no sign of convergence to the StructRL level within the tested horizon. The budgets remain short compared with those of image-based benchmarks, and we make no statement about the asymptotic behaviour of either arm.
\label{sec:longbudget}

\begin{table}[t]
\centering\small
\caption{Three times the training budget. Final: success of the last evaluation checkpoint; Tail: mean success of the last five of 25 checkpoints, both averaged over the listed seeds (500--509); columns $1\times$ give the main block at its budget for the same seeds, columns $3\times$ the runs with three times the budget. One deterministic evaluation episode per checkpoint; paired bootstrap intervals (5000 draws).}
\label{tab:long}
\begin{tabular}{@{}llrrrrrr@{}}
\toprule
Task & Arm & Budget ($1\times$ / $3\times$) & Seeds & Final $1\times$ & Final $3\times$ & Tail $1\times$ & Tail $3\times$ \\
\midrule
DoorKey 6x6 & DQN baseline & 60k / 180k & 10 & 0.10 & 0.30 & 0.04 & 0.28 \\
 & StructRL & 60k / 180k & 10 & 0.90 & 1.00 & 0.86 & 0.92 \\ \multicolumn{8}{r}{\footnotesize paired tail difference StructRL $-$ baseline: $+0.64$ $[+0.36,+0.90]$} \\
KeyCorridor S3R2 & DQN baseline & 150k / 450k & 10 & 0.10 & 0.00 & 0.10 & 0.00 \\
 & StructRL & 150k / 450k & 10 & 0.80 & 0.80 & 0.74 & 0.92 \\ \multicolumn{8}{r}{\footnotesize paired tail difference StructRL $-$ baseline: $+0.92$ $[+0.86,+0.98]$} \\
ObstructedMaze 1Dl & DQN baseline & 100k / 300k & 10 & 0.00 & 0.00 & 0.00 & 0.00 \\
 & StructRL & 100k / 300k & 10 & 0.50 & 0.60 & 0.46 & 0.62 \\ \multicolumn{8}{r}{\footnotesize paired tail difference StructRL $-$ baseline: $+0.62$ $[+0.36,+0.86]$} \\
\bottomrule
\end{tabular}
\end{table}

\begin{figure}[t]
\centering
\includegraphics[width=0.88\linewidth]{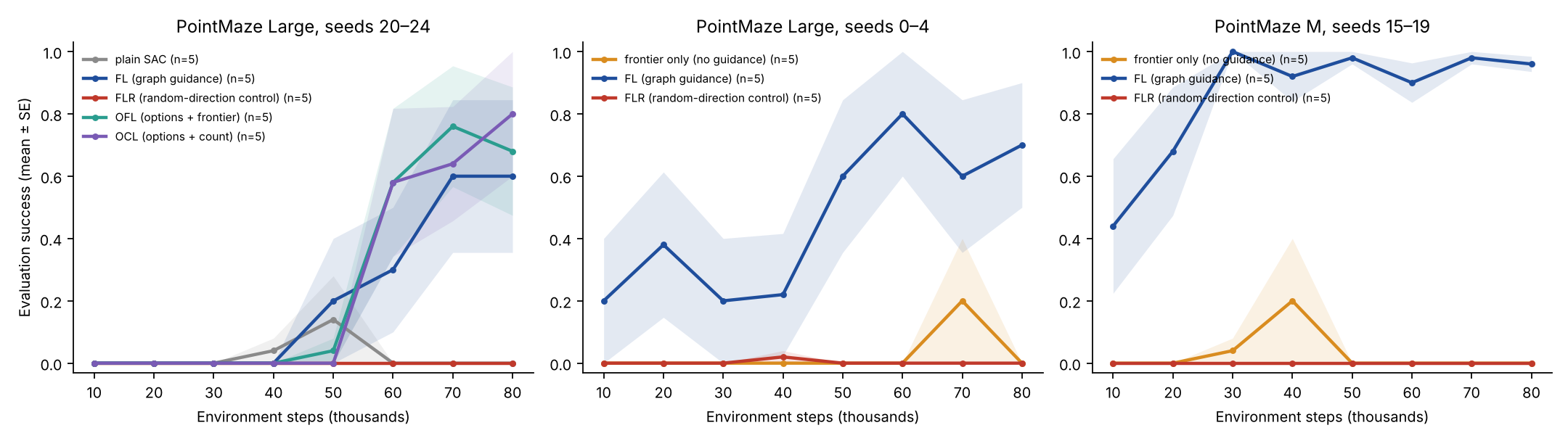}
\caption{Raw, unsmoothed SAC evaluation success over environment steps. StructRL-SAC uses graph-derived actor labels after the anchor; the random-label control has identical pre-anchor exploration but random post-anchor directions.}
\label{fig:sac-curves}
\end{figure}

\paragraph{SAC on PointMaze.} Do graph-derived direction labels improve SAC? On PointMaze Large the label and its random-label control share the same pre-anchor stage, so they differ only in the post-anchor direction label. On the seven runs of the known-anchor block in which the matched arms reach the anchor, final success is 1.0 with graph labels and 0.17 without them, and the paired post-anchor AUC difference is $+0.66$ $[+0.46,+0.83]$ (Figure~\ref{fig:sac-curves}). The exploration prior shows no reliable effect on SAC, and we do not interpret it as evidence that a known anchor direction accelerates search.

\paragraph{Final controller.} Does the graph policy yield a controller that succeeds from many states? On DoorKey $6\times6$ the distilled StructRL+Transfer policy reaches the goal from 0.91 of the reachable states under random-action noise, against 0.38 for plain DQN; ordinary StructRL reaches 0.90, and DoorKey $8\times8$ shows the same ordering (0.86 against 0.00 for plain DQN on a fixed sample of 600 states).

\subsection{Why does it work?}
\paragraph{Graph-derived action preferences.} Is the gain due to the graph actions or to the auxiliary loss? On ten new DoorKey seeds, adding action-preference transfer raises mean AUC from 0.456 to 0.652 ($+0.196$ $[+0.084,+0.308]$), and shuffling the optimal action sets within anchor-distance shells reduces it to 0.000. The gain therefore depends on the graph-derived preference. It extends to the held-out tasks UnlockPickup and KeyCorridor S3R3 (Sections~\ref{sec:dqn-transfer} and \ref{sec:heldout} in Appendix~\ref{app:details}).

\paragraph{Graph policy and Q-model.} Where is the graph policy stronger than the learned model? On the supported states the executed graph controller $\pi_{\mathrm{exec}}$ reaches a mean AUC of 0.95, against 0.46 for the Q-model of the same runs (ordinary StructRL), so the transfer is incomplete. Off the support the Q-model is worse than a random action (Section~\ref{sec:hybrid}).

\paragraph{Distillation.} What does the final distillation phase add? Hard distillation raises the coverage of StructRL+Transfer from 0.84 to 0.91, and graph-derived labels outperform random supported-action labels by $+0.671$ $[+0.580,+0.761]$ (Section~\ref{sec:coverage}).

\paragraph{Comparison with TER.} Does a graph-ordered replay give the same gain? In our DQN implementation, Topological Experience Replay produces no detectable AUC gain over uniform replay on DoorKey $6\times6$ ($+0.08$ $[-0.06,+0.22]$), and StructRL with margin and prior is better on all three tasks (Section~\ref{sec:ter}). The comparison uses symbolic observations and budgets of 60k to 120k steps, and it does not show that TER is ineffective in its original setting.

\subsection{Where does it fail?}
\label{sec:scaling}
The 99-run scaling series (Table~\ref{tab:scaling}) varies task size, structural representation and nuisance observation dimension separately; the axes are not pooled.

\paragraph{Support: larger tasks.} On DoorKey $8\times8$ the prior-plus-margin arm remains effective ($+0.52$ AUC $[+0.32,+0.68]$ over plain DQN), but on $16\times16$ the Q-model learns nothing within the budget although the graph policy succeeds in 2/3 seeds: useful graph support is not turned into a learned policy. This is not evidence for scaling beyond $8\times8$.

\paragraph{Abstraction: coarse structural states.} Removing the heading or keeping only position and carried object reduces AUC from 0.69 to 0.00 in all ten seeds, and graph-only success falls to 0.00. Along the optimal path 39\% and 50\% of the merged classes require conflicting actions, so a policy based only on these aggregated states cannot select an optimal action for every underlying state.

\paragraph{Transfer: high-dimensional inputs.} With 1000 nuisance dimensions the joint Q-model fails in all ten seeds although the graph reaches the anchor in all ten and the graph-only policy succeeds with rate 1.0; a two-branch Q-model that receives $\phi(s)$ directly succeeds in all ten (Section~\ref{sec:featureaccess}). Feature access is not sufficient on the larger task: on DoorKey $8\times8$ removing graph episodes leaves stable success in 3/10 seeds against 10/10 with them ($+0.40$ AUC $[0.28,0.52]$). Graph planning time is not the dominant cost at the tested sizes (at most 117 s per run, graphs of up to about 25k nodes).

\begin{table*}[t]
\caption{Controlled DQN scaling results. The three axes are interpreted separately.}
\label{tab:scaling}
\centering\small
\begin{tabular}{p{2.7cm}p{3.0cm}p{8.2cm}}
\toprule
Axis & Condition & Result\\
\midrule
Task size & DoorKey 8x8 & Prior+margin vs plain: $+0.52$ AUC $[+0.32,+0.68]$; vs StructRL: $+0.34$ $[+0.15,+0.54]$; 10 seeds.\\
Task size & DoorKey 16x16 & 3 seeds: Q-model AUC 0 for all configurations; graph-only success 0.67; no claim beyond 8x8.\\
Structural $\phi$ & \texttt{nodir}, \texttt{pos} & AUC 0.00; graph-only 0.00; optimal-path conflicting classes 39\% and 50\%.\\
Nuisance input & 100 / 1000 dims & At 100: $-0.17$ AUC vs no nuisance; at 1000: AUC 0.008 while graph-only remains 1.0.\\
Planning & up to 25k nodes & Total planning time per run at most 117 s; not the dominant tested bottleneck.\\
\bottomrule
\end{tabular}
\end{table*}

\paragraph{Discovery as a fallback.} When no supported path exists, a correct prior combined with negative evidence reduces the mean search in SearchGrid from 1290 to 970 steps at $15\times15$ and from 51,442 to 18,531 steps at $50\times50$ (Appendix~\ref{app:details}); discovery is a fallback for missing support, not the main performance claim.

\section{Analysis and Discussion}
\label{sec:discussion}

\subsection{Graph planning and learned RL models}
The replay graph and the learned RL models serve different roles. On supported states, anchor-relative dynamic programming gives a strong controller directly from observed transitions. DQN and SAC can absorb part of this action information, but the quality of that transfer depends on the model and input representation. In DQN, the executed graph controller generally remains stronger than the online Q-model on supported states. In SAC, graph direction labels can bring the actor model close to the graph policy once anchor support exists.

The final-policy experiment separates policy quality from coverage: randomization raises both controllers, distillation adds more when the online Q-model is weaker, and random-label distillation fails, so the graph labels remain necessary.

\subsection{StructRL as a multi-stage framework}
The present implementation is one realization of a multi-stage StructRL framework rather than a fully optimized algorithm, and each stage has its own limit. Policy transfer is incomplete: the learned model absorbs only part of the action preferences of the graph policy, and in all three tasks of the TER block the executed graph controller is stronger than the transferred Q-model on the supported states. Structural representation remains a design problem: automatic input selection removes nuisance dimensions but does not reach the performance of the hand-built $\phi$. Support acquisition remains interaction-intensive even when graph-based discovery helps. The experiments thus isolate three bottlenecks: graph-to-model transfer, structural representation and support acquisition. They do not establish how much of the remaining gap can be closed on larger or real systems.

\subsection{Known goals and graph information}
A known goal does not specify a route. In DQN, the Manhattan-distance prior can shorten support acquisition on harder tasks. In SAC, the matched prior experiments do not show a direction-specific search benefit. The result that is consistent across both model families is the benefit of graph-derived action information after usable graph support has been observed.

\subsection{Scaling limits}
The scaling experiments identify three separate failures. First, larger tasks can produce useful graph support that the Q-model does not learn within the training budget. Second, an overly coarse $\phi$ merges states that require different actions and invalidates graph planning. Third, high-dimensional nuisance input can prevent the DQN Q-model from using graph-derived labels even when the graph itself remains correct. A two-branch Q-model that receives $\phi(s)$ directly removes the third failure. The first two remain open. Explicit graph-planning time is not the dominant cost at the tested graph sizes.

\subsection{Limitations}
The evidence is dominated by small symbolic MiniGrid tasks with fixed layouts. DoorKey $8\times8$ and continuous PointMaze broaden the setting but do not establish image-based or large-scale scalability. The main method uses a task-derived $\phi$; the selection rule in Section~\ref{sec:phi-determine} requires a structural key and creates no new features. PointMaze cells are only partly transition-consistent, so the deterministic graph guarantee does not apply there. Several controls use few seeds, and the final-policy experiments are limited to the DoorKey family. Their off-support success does not separate generalization from recovery through random actions. The TER comparison uses the official graph, search and encoder components but not the original training regime. The budget comparison of Section~\ref{sec:longbudget} extends the DQN blocks to three times the budget; longer training of plain SAC was not evaluated. We provide no real-system evaluation, asymptotic scaling result or convergence guarantee beyond the stated abstraction and support propositions. Appendix~\ref{app:protocol} lists the run protocol and the deviations from it.

\section{Conclusion}
StructRL shows that an empirical replay graph can provide action supervision for model-free learners without replacing their learning targets. The experiments separate three limitations: incomplete observed support, structural aliasing, and unsuccessful transfer of graph decisions to the learned model. For DQN, graph guidance improves success on all three primary MiniGrid tasks, and the advantage over plain DQN persists at three times the training budget; separate experiments isolate the additional effect of transferring the action preferences. For SAC, graph-derived direction labels improve the actor on PointMaze after goal contact. The results hold for the tested structured environments with a task-specified $\phi$; automatic construction of $\phi$, stronger transfer, and performance beyond the observed support remain open.

\section*{Use of AI tools}
Large language models (Claude, Anthropic; ChatGPT, OpenAI) assisted with software implementation, the execution of the experiment scripts, data processing, table and figure generation, and the drafting and revision of the manuscript. The author was responsible for the research questions, the method, the experimental design and the interpretation of the results. The reported results are based on stored experiment outputs. Selected numerical results are regenerated by analysis scripts and unit tests check the core graph and loss computations; code and stored results are available from the author on request. The training runs of all tables were not repeated by a second party. The author reviewed the AI-assisted material and is responsible for the accuracy and integrity of the work.

\appendix
\small
\section{Detailed results}
\label{app:details}
The full result paragraphs and tables behind Section~\ref{sec:results}, with intervals and sample sizes. The appendix documents the results; the argument is in the main text.

\subsection{Explicit graph structure improves DQN performance}
Table~\ref{tab:dqn-night} gives the main DQN results. StructRL reaches the anchor in 10/10 seeds on all three tasks and attains mean evaluation AUCs of 0.41, 0.44 and 0.23 on DoorKey, KeyCorridor S3R2 and ObstructedMaze. Plain DQN reaches the anchor in 6/10, 4/10 and 1/10 seeds and attains AUCs of 0.01, 0.04 and 0.00. Discovery-only StructRL reaches the anchor in 10/10 seeds on every task but has AUC 0.00, separating reliable anchor acquisition from the post-anchor performance mechanism.

\begin{table}[t]
\centering\small
\caption{\textbf{Seeds 500--509.} Anchor found (of 10 seeds), mean evaluation AUC, mean final evaluation success, and paired difference in AUC (StructRL minus arm; bootstrap 95\% interval, 5000 draws). One deterministic evaluation episode per checkpoint; no multiplicity correction.}
\label{tab:dqn-night}
\begin{tabular}{@{}llrrrl@{}}
\toprule
Task & Arm & Anchor & AUC & Final & $\Delta$AUC (StructRL $-$ arm) \\
\midrule
DoorKey 6x6 & DQN baseline & 6/10 & 0.01 & 0.10 & +0.40 [+0.29,+0.50] \\
 & count-only control & 9/10 & 0.04 & 0.10 & +0.36 [+0.27,+0.46] \\
 & random-direction control & 8/10 & 0.05 & 0.20 & +0.36 [+0.27,+0.45] \\
 & discovery-only StructRL & 10/10 & 0.00 & 0.00 & +0.40 [+0.30,+0.50] \\
 & StructRL & 10/10 & 0.41 & 0.90 & -- \\
\midrule
KeyCorridor S3R2 & DQN baseline & 4/10 & 0.04 & 0.10 & +0.40 [+0.25,+0.54] \\
 & count-only control & 8/10 & 0.10 & 0.10 & +0.34 [+0.18,+0.48] \\
 & random-direction control & 4/10 & 0.06 & 0.10 & +0.38 [+0.22,+0.53] \\
 & discovery-only StructRL & 10/10 & 0.00 & 0.00 & +0.44 [+0.28,+0.58] \\
 & StructRL & 10/10 & 0.44 & 0.80 & -- \\
\midrule
ObstructedMaze & DQN baseline & 1/10 & 0.00 & 0.00 & +0.23 [+0.10,+0.38] \\
 & count-only control & 3/10 & 0.00 & 0.00 & +0.23 [+0.08,+0.38] \\
 & random-direction control & 3/10 & 0.00 & 0.00 & +0.23 [+0.08,+0.39] \\
 & discovery-only StructRL & 10/10 & 0.00 & 0.00 & +0.23 [+0.08,+0.39] \\
 & StructRL & 10/10 & 0.23 & 0.50 & -- \\
\bottomrule
\end{tabular}
\end{table}

Figure~\ref{fig:dqn-curves} shows the DQN curves without smoothing. Each checkpoint contains one deterministic evaluation episode per seed, so the mean curve is coarse and can visibly step as individual seeds switch between failure and success.

\subsection{Graph-derived action preferences transfer into DQN}
\label{sec:dqn-transfer}
The DoorKey transfer test uses ten new seeds and fixes the margin and weight before the first run. Mean AUC increases from 0.456 for StructRL to 0.652 for StructRL with graph-action transfer, a paired difference of $+0.196$ with bootstrap 95\% interval $[+0.084,+0.308]$. With graph-optimal action sets shuffled within anchor-distance shells, mean AUC is 0.000. This control shows that the gain depends on the graph-derived action preference rather than on adding an auxiliary loss alone.

In the 10-seed known-anchor block, adding the margin to the matched prior-only arm changes AUC by $+0.23$ $[+0.18,+0.30]$ on DoorKey, $+0.06$ $[+0.02,+0.10]$ on KeyCorridor S3R2, and $+0.20$ $[+0.06,+0.35]$ on ObstructedMaze. The KeyCorridor increment is small but its paired interval remains above zero. The margin transfers action rankings rather than calibrated graph values.

\subsection{DQN transfer on held-out tasks}
\label{sec:heldout}
On UnlockPickup, the prior-only arm improves AUC over ordinary StructRL by $+0.22$ $[+0.12,+0.34]$ and shortens anchor acquisition by about 23k steps $[-31\text{k},-16\text{k}]$. Adding the margin beyond the prior contributes another $+0.19$ $[+0.08,+0.30]$, for a total gain of $+0.42$ $[+0.24,+0.58]$ over StructRL. On held-out KeyCorridor S3R3, the prior effect is marginal at $+0.06$ $[0.00,+0.14]$, the additional margin contrast is $+0.05$ with an interval touching zero, and the combined prior-plus-margin arm improves AUC over StructRL by $+0.11$ $[+0.02,+0.21]$. Final success on KeyCorridor S3R3 is 0.40 for prior-plus-margin versus 0.20 for StructRL and prior-only. Plain DQN reaches the anchor in 1/10 UnlockPickup seeds and 0/10 KeyCorridor S3R3 seeds.

\begin{table}[t]
\caption{Held-out DQN known-anchor results, paired AUC differences.}
\label{tab:heldout-known}
\centering\small
\begin{tabular}{lccc}
\toprule
Task & Prior $-$ StructRL & Margin beyond prior & Prior+Margin $-$ StructRL\\
\midrule
UnlockPickup & $+0.22$ $[0.12,0.34]$ & $+0.19$ $[0.08,0.30]$ & $+0.42$ $[0.24,0.58]$\\
KeyCorridor S3R3 & $+0.06$ $[0.00,0.14]$ & $+0.05$ $[-0.00,0.13]$ & $+0.11$ $[0.02,0.21]$\\
\bottomrule
\end{tabular}
\end{table}

\subsection{Graph direction labels improve SAC}
On PointMaze Large, StructRL-SAC and its random-label control share exactly the same pre-anchor stage and reach the anchor in the same three of five seeds. They differ only in the post-anchor direction label. In those three seeds StructRL-SAC reaches post-anchor training-success AUCs of 0.78--0.89 and final evaluation success 1.0, whereas the random-label control remains at 0.02--0.05 with final success 0. Figure~\ref{fig:sac-curves} shows the curves. The deterministic evaluation uses the SAC actor itself; graph exploration and graph actions are not used during evaluation.

\subsection{Graph labels improve SAC after anchor contact}
\label{sec:sac-known}
The PointMaze Large known-anchor block (Table~\ref{tab:sac-known}) contains 50 runs (seeds 600--609, five arms). The primary transfer contrast is strong: on the seven seeds where the matched arms reach the anchor, the graph-label minus no-label contrast ($FL_p-F_p$) in post-anchor AUC is $+0.66$ $[+0.46,+0.83]$, positive in all seven seeds. On these same seeds, The matched StructRL-SAC model with graph labels reaches final success 1.0 in 7/7, while the goal-prior model without graph labels has mean final success 0.17. Graph-only success for the no-label runs is 0.41 on average, indicating that useful graph structure is present but reaches the actor poorly without labels.

The exploration-prior result is different. Relative to ordinary StructRL-SAC, the goal-prior StructRL-SAC model reaches the anchor 13.7k steps earlier on average $[-25.8\text{k},-3.2\text{k}]$, but the wrong-prior arm reaches the anchor more often (9/10 versus 7/10) and earlier in median terms (40k versus 48k). The direct correct-versus-wrong contrast has the opposite sign to the directional hypothesis and an interval spanning zero. We therefore do not interpret the SAC prior result as evidence that known anchor direction accelerates search. The overall post-anchor contrast between goal-prior StructRL-SAC and ordinary StructRL-SAC is also not different from zero. Against the SAC baseline, however, the goal-prior StructRL-SAC model improves evaluation AUC by $+0.28$ $[+0.15,+0.41]$.

\begin{table}[t]
\caption{SAC known-anchor comparisons on PointMaze Large.}
\label{tab:sac-known}
\centering\small
\begin{tabular}{lcc}
\toprule
Contrast & Mean difference [95\% CI] & $n$\\
\midrule
Transfer: $FL_p-F_p$ post-anchor AUC & $+0.661$ $[+0.464,+0.832]$ & 7\\
Prior: $T_{anchor}(FL_p-FL)$ & $-13.7$k $[-25.8,-3.2]$k & 10\\
Direction: $T_{anchor}(FL_p-FL_{pwrong})$ & $+9.3$k $[-8.3,+26.8]$k & 10\\
Overall post-anchor: $FL_p-FL$ & $-0.077$ $[-0.354,+0.222]$ & 6\\
Against plain SAC: eval AUC & $+0.280$ $[+0.146,+0.411]$ & 10\\
\bottomrule
\end{tabular}
\end{table}

\subsection{Scaling reveals three bottlenecks}
The 99-run scaling series (Table~\ref{tab:scaling}) varies task size, structural representation and nuisance observation dimension separately. We report these axes separately and do not pool across tasks.

\paragraph{Task size.}
From DoorKey $6\times6$ to $8\times8$, the prior-plus-margin arm remains effective: relative to plain DQN the paired AUC gain is $+0.52$ $[+0.32,+0.68]$, and relative to ordinary StructRL it is $+0.34$ $[+0.15,+0.54]$ across ten seeds. Plain DQN reaches the anchor in 6/10 seeds and StructRL in 10/10. On DoorKey $16\times16$, only three seeds are available. Plain DQN reaches no anchor; the prior arms reach it in 2/3 seeds, but evaluation AUC remains zero for all arms within the budget. The graph-only policy nevertheless succeeds in 2/3 seeds (success 0.67), so this run shows usable graph structure without successful Q-model learning. This is not evidence for scaling beyond $8\times8$.

\paragraph{Structural representation.}
Removing heading from $\phi$ (\texttt{nodir}) or retaining only position and carried object (\texttt{pos}) reduces evaluation AUC to 0.00 in all ten seeds, compared with 0.69 for the exact structural representation. Anchor acquisition still occurs in 9/10 seeds, but graph-only success also falls to 0.00. A separate exact-task aliasing analysis accounts for this failure: along the optimal path, 39\% of \texttt{nodir} classes and 50\% of \texttt{pos} classes merge states that require conflicting best actions. The corresponding upper bounds on per-state action correctness are about 74\% and 65\%; the exact representation has no such conflicts on the path. The analysis is diagnostic and is measured only on DoorKey $6\times6$.

\paragraph{Observation dimension.}
With 100 independently resampled nuisance dimensions, the prior-plus-margin arm remains well above plain DQN but loses 0.17 AUC $[-0.31,-0.05]$ relative to the no-nuisance condition. With 1000 nuisance dimensions, its AUC falls to 0.008, essentially plain-DQN performance, while the graph-only policy still succeeds with rate 1.0. Thus the structural graph remains useful but graph-derived supervision no longer transfers effectively into the raw-observation Q-model at this input dimension. Section~\ref{sec:featureaccess} shows that this failure is removed when the Q-model receives the structural features through a separate branch.

\paragraph{Planning cost.}
Across the scaling series, explicit graph planning remains inexpensive relative to environment interaction: total planning time per run ranges from seconds to about two minutes (maximum 117 s) for runs of 3 to 40 minutes, with graphs of up to about 25k nodes. Planning time depends on the number of replanning calls as well as on graph size. This does not establish asymptotic scalability, but planning is not the dominant bottleneck at the tested sizes.

\subsection{Feature access and graph necessity}
\label{sec:featureaccess}
\begin{table*}[t]
\centering\small
\caption{Feature access under nuisance input (DoorKey, StructRL with goal prior and graph-action transfer). Stable = first of two consecutive evaluation checkpoints with success 1; final = success at the last checkpoint; relapses = success $1\to0$ transitions summed over seeds; AUC = mean evaluation success over 25 checkpoints. Seeds 1710--1719 (confirmation), 1720--1724 (F2), 1730--1734 (F3).}
\label{tab:featureaccess}
\begin{tabular}{@{}llcccrr@{}}
\toprule
Configuration & Q-model input & $n$ & Stable & Final $=1$ & AUC median [min,max] & Relapses\\
\midrule
\multicolumn{7}{@{}l}{\textit{DoorKey 6x6, 1000 nuisance dimensions, 60k steps (confirmation)}}\\
Full StructRL & joint Q-model & 10 & 0/10 & 0/10 & 0.00 [0.00,0.00] & 0\\
Full StructRL & two-branch Q-model & 10 & 10/10 & 9/10 & 0.80 [0.44,0.96] & 18\\
Full StructRL & $\phi$-only Q-model & 10 & 10/10 & 9/10 & 0.84 [0.52,0.96] & 13\\
\midrule
\multicolumn{7}{@{}l}{\textit{DoorKey 6x6, no nuisance (F2)}}\\
Full StructRL & joint Q-model & 5 & 5/5 & 5/5 & 0.64 [0.52,0.80] & 0\\
Full StructRL & two-branch Q-model & 5 & 5/5 & 5/5 & 0.92 [0.80,0.96] & 0\\
\midrule
\multicolumn{7}{@{}l}{\textit{DoorKey 6x6, 100 nuisance dimensions (F2)}}\\
Full StructRL & joint Q-model & 5 & 5/5 & 4/5 & 0.48 [0.08,0.64] & 4\\
Full StructRL & two-branch Q-model & 5 & 5/5 & 5/5 & 0.84 [0.80,0.92] & 2\\
\midrule
\multicolumn{7}{@{}l}{\textit{DoorKey 6x6, 5000 nuisance dimensions (F2)}}\\
Full StructRL & joint Q-model & 5 & 0/5 & 0/5 & 0.00 [0.00,0.04] & 1\\
Full StructRL & two-branch Q-model & 5 & 5/5 & 4/5 & 0.76 [0.60,0.88] & 7\\
\midrule
\multicolumn{7}{@{}l}{\textit{DoorKey 8x8, 1000 nuisance dimensions (F3)}}\\
Full StructRL & joint Q-model & 5 & 0/5 & 0/5 & 0.00 [0.00,0.00] & 0\\
Full StructRL & two-branch Q-model & 5 & 5/5 & 4/5 & 0.84 [0.36,0.88] & 11\\
\bottomrule
\end{tabular}
\end{table*}

With 1000 nuisance dimensions on DoorKey $6\times6$, the joint Q-model fails in all ten seeds although the graph reaches the anchor in all ten; the two-branch Q-model succeeds stably in all ten. The same qualitative separation holds at 5000 nuisance dimensions and on DoorKey $8\times8$ (Table~\ref{tab:featureaccess}). This identifies access to the structural features, rather than failure of the graph itself, as the main bottleneck in this setting.

\begin{table*}[t]
\centering\small
\caption{Graph necessity beyond feature access (two-branch Q-model in all configurations; seeds 1780--1784 on DoorKey 6x6 and 1790--1799 on DoorKey 8x8, 1000 nuisance dimensions; the anchor is known only in the full StructRL configuration, in the other arms it is the reward cell found by the agent. DQN baseline: no graph episodes or margin; StructRL: graph episodes without margin; Full StructRL: goal prior plus graph-action transfer. Columns as in Table~\ref{tab:featureaccess}.}
\label{tab:necessity}
\begin{tabular}{@{}llcccrr@{}}
\toprule
Configuration & Q-model input & $n$ & Stable & Final $=1$ & AUC median [min,max] & Relapses\\
\midrule
\multicolumn{7}{@{}l}{\textit{DoorKey 6x6, 1000 nuisance dimensions}}\\
DQN baseline & two-branch Q-model & 5 & 3/5 & 4/5 & 0.24 [0.00,0.44] & 2\\
StructRL & two-branch Q-model & 5 & 5/5 & 3/5 & 0.72 [0.56,0.92] & 8\\
Full StructRL & two-branch Q-model & 5 & 5/5 & 4/5 & 0.76 [0.48,0.80] & 20\\
\midrule
\multicolumn{7}{@{}l}{\textit{DoorKey 8x8, 1000 nuisance dimensions}}\\
DQN baseline & two-branch Q-model & 10 & 3/10 & 1/10 & 0.00 [0.00,0.52] & 6\\
StructRL & two-branch Q-model & 10 & 10/10 & 7/10 & 0.48 [0.24,0.80] & 22\\
Full StructRL & two-branch Q-model & 10 & 10/10 & 6/10 & 0.46 [0.32,0.80] & 41\\
\bottomrule
\end{tabular}
\end{table*}

Feature access is not sufficient on the larger task. With the two-branch architecture fixed, removing graph episodes leaves DoorKey $8\times8$ with stable success in 3/10 seeds, whereas graph episodes without the margin succeed stably in 10/10; the paired AUC difference is $+0.40$ with a 95\% bootstrap interval of $[0.28,0.52]$ (Table~\ref{tab:necessity}). On $6\times6$ the graph mainly accelerates learning. In this configuration the margin adds no further AUC benefit beyond graph episodes, while the known-anchor prior advances first goal contact.

Two controls do not support gradient routing or an explicit supervised DP head as the explanation. Detaching the raw branch in the graph/margin step leaves the two-branch result unchanged (median AUC 0.88 with and without detaching, five seeds), and the gradient cosines do not show a larger conflict in the failing joint Q-model; an explicit dynamic-programming head does not improve the two-branch Q-model (median AUC 0.64 against 0.64, and 0.76 with confidence weighting).

\subsection{Comparison with Topological Experience Replay}
\label{sec:ter}
Table~\ref{tab:ter} gives the TER comparison on DoorKey $6\times6$, UnlockPickup and DoorKey $8\times8$. On DoorKey $6\times6$, TER produces no detectable AUC gain over uniform replay, neither for plain DQN (D1, $+0.08$ $[-0.06,+0.22]$) nor on top of graph episodes (D2, $+0.03$ $[-0.14,+0.21]$). On UnlockPickup and DoorKey $8\times8$ the plain arms stay at AUC 0.00, so D1 is uninformative there, and D2 is again not distinguishable from zero ($+0.10$ $[0.00,+0.23]$ and $+0.04$ $[-0.17,+0.24]$). StructRL with margin and known-anchor prior is better than DQN with TER on all three tasks (D4, $+0.58$ $[+0.47,+0.69]$, $+0.30$ $[+0.08,+0.50]$ and $+0.35$ $[+0.11,+0.59]$), and so is the adaptation of the learner to the graph (D5 $>0$ on all three); graph episodes alone (D3) are clearly better only on DoorKey $6\times6$ ($+0.32$ $[+0.22,+0.42]$). In our DQN implementation and these settings, TER therefore shows no detectable AUC gain over uniform replay. The comparison deviates from the original TER setup in five respects: the non-TER half of each minibatch is drawn uniformly instead of by prioritized replay; the learner is our Double+Dueling+$n$-step DQN with its $\epsilon$ schedule; the observation is the symbolic full-grid encoding instead of an RGB image; the budget is 60k to 120k steps instead of millions; and TER is not applied to SAC, since it is a replay-ordering method for Q-learning. The result concerns this configuration and does not show that TER is ineffective in its original setting. It is consistent with the different points of intervention: TER reorders backups on data that have already been collected, whereas StructRL changes which data are collected and which actions the learner prefers.

\begin{table}[t]
\caption{TER comparison on DoorKey $6\times6$ (seeds 1100--1109, 60k steps), UnlockPickup (seeds 1200--1204, 100k steps) and DoorKey $8\times8$ (seeds 1300--1304, 120k steps). Top: Q-model evaluation AUC. Internal arm codes are retained here because the contrasts use these names: \texttt{plain}=DQN baseline, \texttt{plain\_tero}=TER baseline, \texttt{struct}=StructRL, \texttt{struct\_tero}=StructRL with TER replay, and \texttt{struct\_mgkahg}=StructRL with goal prior and graph-action transfer. Top: Q-model evaluation AUC (runs in which the anchor was found). Bottom: paired differences, mean with bootstrap 95\% interval (5000 draws). D6 compares the graph policy executed alone with the Q-model of the same run at the same checkpoints. On UnlockPickup and DoorKey $8\times8$ both plain arms are at AUC 0.00, so D1 is uninformative there.}
\label{tab:ter}
\centering\footnotesize
\begin{tabular}{lccc}
\toprule
Arm & DoorKey $6\times6$ & UnlockPickup & DoorKey $8\times8$ \\
\midrule
\texttt{plain} & 0.06 (7/10) & 0.00 (2/5) & 0.00 (3/5) \\
\texttt{plain\_tero} & 0.13 (8/10) & 0.00 (0/5) & 0.00 (3/5) \\
\texttt{struct} & 0.46 (10/10) & 0.06 (5/5) & 0.06 (5/5) \\
\texttt{struct\_tero} & 0.49 (10/10) & 0.17 (5/5) & 0.10 (5/5) \\
\texttt{struct\_mgkahg} & 0.72 (10/10) & 0.30 (5/5) & 0.35 (5/5) \\
\midrule
Contrast & $n=10$ & $n=5$ & $n=5$ \\
\midrule
D1 \texttt{plain\_tero} $-$ \texttt{plain} & $+0.08$ $[-0.06,+0.22]$ & $0.00$ $[0.00,0.00]$ & $0.00$ $[0.00,0.00]$ \\
D2 \texttt{struct\_tero} $-$ \texttt{struct} & $+0.03$ $[-0.14,+0.21]$ & $+0.10$ $[0.00,+0.23]$ & $+0.04$ $[-0.17,+0.24]$ \\
D3 \texttt{struct} $-$ \texttt{plain\_tero} & $+0.32$ $[+0.22,+0.42]$ & $+0.06$ $[0.00,+0.19]$ & $+0.06$ $[0.00,+0.19]$ \\
D4 \texttt{struct\_mgkahg} $-$ \texttt{plain\_tero} & $+0.58$ $[+0.47,+0.69]$ & $+0.30$ $[+0.08,+0.50]$ & $+0.35$ $[+0.11,+0.59]$ \\
D5 \texttt{struct\_mgkahg} $-$ \texttt{struct} & $+0.26$ $[+0.16,+0.37]$ & $+0.23$ $[+0.06,+0.40]$ & $+0.29$ $[+0.10,+0.48]$ \\
D6 graph $-$ Q-model, \texttt{struct} & $+0.49$ $[+0.40,+0.60]$ & $+0.70$ $[+0.60,+0.82]$ & $+0.75$ $[+0.55,+0.91]$ \\
\hphantom{D6 }(graph vs Q-model AUC) & 0.95 vs 0.46 & 0.76 vs 0.06 & 0.82 vs 0.06 \\
D6 graph $-$ Q-model, \texttt{struct\_mgkahg} & $+0.26$ $[+0.22,+0.30]$ & $+0.62$ $[+0.44,+0.79]$ & $+0.57$ $[+0.37,+0.77]$ \\
\hphantom{D6 }(graph vs Q-model AUC) & 0.98 vs 0.72 & 0.91 vs 0.30 & 0.92 vs 0.35 \\
\bottomrule
\end{tabular}
\end{table}

\subsection{Graph policy and Q-model on and off the observed support}
\label{sec:hybrid}
\paragraph{On the support the executed graph controller is stronger.}
Within the TER block, the executed graph controller $\pi_{\mathrm{exec}}$ without the Q-model reaches a mean AUC of 0.95 for ordinary StructRL and 0.98 for StructRL+Transfer, against 0.46 and 0.72 for the Q-models of the same runs (D6, $+0.49$ $[+0.40,+0.60]$ and $+0.26$ $[+0.22,+0.30]$; Table~\ref{tab:ter}) on DoorKey $6\times6$; on UnlockPickup and DoorKey $8\times8$ $\pi_{\mathrm{exec}}$ is likewise ahead of the Q-model (D6 between $+0.57$ and $+0.75$). Within the given training budget, anchor-relative dynamic programming on the empirical replay graph therefore yields the stronger policy on the observed support. Adapting the learner to the graph, by adding the margin and the known-anchor prior to the graph episodes, raises the Q-model AUC by $+0.26$ $[+0.16,+0.37]$ (D5) and roughly halves the gap to $\pi_{\mathrm{exec}}$ without closing it. D5 does not separate margin and prior; the margin beyond a matched prior is positive on DoorKey in the known-anchor block (Section~\ref{sec:dqn-transfer}; Table~\ref{tab:heldout-known} for the held-out tasks).

\paragraph{Off the support the Q-model is worse than a random action.}
In the hybrid block, start states are perturbed by random-action prefixes of up to 400 steps. From starts on the support, the graph controllers succeed in 0.96 (ordinary StructRL) and 0.95 (StructRL+Transfer) of the episodes and the Q-model alone in 0.40 and 0.90; the adapted Q-model thus follows $\pi_{\mathrm{exec}}$ closely on the support. Off-support starts are rare (5\% and 9\% of all starts). Where the graph has no supported action, continuing with the Q-model instead of a uniformly random action lowers success: $\Delta_{\mathrm{off}}=-0.39$ $[-0.59,-0.23]$ for ordinary StructRL (8 seeds with at least five off-support starts) and $-0.24$ $[-0.40,-0.12]$ for StructRL+Transfer (9 seeds). The graph-chosen steps have $C_\phi=1.0$ in all runs. Within this setting the Q-model does not extend the graph policy beyond its support, and we make no generalization claim.

\paragraph{PointMaze and off-support boundary.} With $C_\phi=0.59$ for the SAC representation, Proposition~\ref{prop:support} does not apply on PointMaze. Additional random-start and larger-maze diagnostics do not establish performance beyond observed support: only four of ten PointMaze Large random-start runs reach the goal, and neither of two runs on a larger $13\times17$ maze reaches it. Full projection comparisons, perturbed-start controls and decision rules are reported in Appendix~\ref{app:boundary}.

\subsection{End-to-end policy comparison, distillation and randomization}
\label{sec:coverage}
The final-policy experiment asks whether the graph-derived policy can be converted into a controller that succeeds from many reachable states, not only from the nominal task start. On DoorKey $6\times6$, with the same random-action noise applied to both controllers, the final StructRL+Transfer policy reaches the goal from 0.91 of reachable states compared with 0.38 for plain DQN. Ordinary StructRL reaches 0.90. Plain DQN with this randomization is not above uniform random actions, so random actions alone do not explain the StructRL result.

The mechanism controls separate graph information, distillation and randomization. For StructRL+Transfer, the Q-model trained during interaction reaches 0.84 coverage under the same randomization, and hard distillation adds approximately $+0.075$ to reach 0.91; graph-derived labels outperform random supported-action labels by $+0.671$ $[+0.580,+0.761]$. For ordinary StructRL, whose online Q-model is weaker, distillation adds about $+0.30$ at equal randomization. Thus the contribution of distillation depends on how much of the graph policy the online Q-model has already learned, while the graph-derived labels are essential in both cases.

DoorKey $8\times8$ shows the same ordering on a fixed sample of 600 reachable states: the final controller reaches 0.86 coverage, the online Q-model 0.22 and plain DQN 0.00. This larger-scale block does not contain the matched-randomization control used on DoorKey $6\times6$. Full controller tables, path lengths, support-aware randomization, soft-target controls, E1/E2 thresholds and bootstrap intervals are reported in Appendix~\ref{app:coverage-details}.

\subsection{Discovery remains a fallback for missing support}
When no supported path is available, graph-based exploration can reduce the amount of search needed to create useful support. In SearchGrid with a tight step budget (runs without an anchor are counted at the budget), a correct prior combined with negative evidence reduces the mean search from 1290 to 970 steps at $15\times15$ (anchor in 9/10 versus 5/10 runs) and from 51,442 to 18,531 steps at $50\times50$ (10/10 versus 5/10). These experiments motivate discovery as a fallback for missing support rather than as the main performance claim.

\section{Final-policy phase: setup and coverage results}
\label{app:final-details}
\label{app:coverage-details}
\paragraph{Setup.} For hard distillation the frozen graph policy provides the set of supported actions that maximize $Q_G(z,a)$ at each labelled structural state; ties are retained. The student Q-model is trained on replay observations whose structural state has a graph label, with 20,000 Adam steps, learning rate $5\cdot10^{-4}$ and batch size 128. The soft-target control uses a Boltzmann distribution over supported graph actions with temperature $\tau=0.02$. The final controller executes the greedy student action with probability $1-\varepsilon$ and a uniformly random action otherwise; the support-aware control randomizes only where the graph has no supported action. These controls use no new reward or environment interaction. Table~\ref{tab:cov} gives the coverage of all controllers on DoorKey $6\times6$, Table~\ref{tab:covpair} the paired differences, Table~\ref{tab:cov8} the corresponding results on DoorKey $8\times8$, and Table~\ref{tab:eps} the randomization control. Over the ten seeds (2000--2009) the share of supported states is $C_G=0.39$ (range $0.32$--$0.48$) for ordinary StructRL and $0.38$ ($0.28$--$0.45$) for StructRL+Transfer, and the share of anchor-reachable states is $C_A=0.35$ ($0.28$--$0.41$) and $0.34$ ($0.25$--$0.42$); about 89\% and 90\% of the supported states are anchor-reachable.
\begin{table}[t]
\caption{Policy coverage on DoorKey $6\times6$ (60k steps, no expansion; seeds 2000--2009; StructRL $n=10$, StructRL with goal prior and graph-action transfer $n=10$). $C_\theta$: share of all breadth-first reachable states from which one deterministic-start episode with a fresh step budget reaches the goal, mean $\pm$ standard deviation over seeds. Steps: mean number of steps to the goal on the states solved by all six controllers of the common set (graph + random fallback and the hard distilled controllers; the other rows are marked --).}
\label{tab:cov}
\centering\footnotesize
\resizebox{\linewidth}{!}{\begin{tabular}{lcccc}
\toprule
 & \multicolumn{2}{c}{StructRL} & \multicolumn{2}{c}{StructRL + goal prior + graph transfer} \\
Controller & $C_\theta$ & Steps & $C_\theta$ & Steps \\
\midrule
Online Q-model & 0.11 $\pm$ 0.05 & -- & 0.45 $\pm$ 0.10 & -- \\
Hard distilled, $\varepsilon=0$ & 0.61 $\pm$ 0.08 & 11.3 & 0.59 $\pm$ 0.09 & 10.9 \\
Hard distilled, $\varepsilon=0.1$ & 0.84 $\pm$ 0.07 & 13.3 & 0.84 $\pm$ 0.08 & 12.8 \\
Hard distilled, $\varepsilon=0.3$ & 0.90 $\pm$ 0.07 & 18.7 & 0.91 $\pm$ 0.06 & 18.2 \\
Hard distilled, $\varepsilon_{\mathrm{off}}=0.1$ & 0.81 $\pm$ 0.07 & 11.7 & 0.81 $\pm$ 0.09 & 11.4 \\
Hard distilled, $\varepsilon_{\mathrm{off}}=0.3$ & 0.86 $\pm$ 0.06 & 12.4 & 0.88 $\pm$ 0.07 & 11.9 \\
Soft distilled, $\varepsilon=0$ & 0.32 $\pm$ 0.03 & -- & 0.34 $\pm$ 0.06 & -- \\
Soft distilled, $\varepsilon_{\mathrm{off}}=0.3$ & 0.63 $\pm$ 0.10 & -- & 0.67 $\pm$ 0.09 & -- \\
Graph + random fallback & 0.82 $\pm$ 0.04 & 35.8 & 0.83 $\pm$ 0.06 & 34.5 \\
\bottomrule
\end{tabular}}
\end{table}

\begin{table}[t]
\caption{Paired differences in $C_\theta$ (same seeds), mean with bootstrap 95\% interval (5000 draws); in brackets the number of seeds with a positive difference. Displayed values are rounded; the decisions use the unrounded values.}
\label{tab:covpair}
\centering\footnotesize
\resizebox{\linewidth}{!}{\begin{tabular}{lcc}
\toprule
Difference & StructRL & StructRL + goal prior + graph transfer \\
\midrule
Hard distilled ($\varepsilon=0$) $-$ online Q-model & $+0.495$ $[+0.440,+0.549]$ (10/10) & $+0.146$ $[+0.098,+0.188]$ (9/10) \\
$\varepsilon=0.3$ $-$ $\varepsilon=0$ (hard distilled) & $+0.293$ $[+0.251,+0.333]$ (10/10) & $+0.318$ $[+0.289,+0.346]$ (10/10) \\
$\varepsilon=0.1$ $-$ $\varepsilon=0$ (hard distilled) & $+0.228$ $[+0.199,+0.259]$ (10/10) & $+0.244$ $[+0.222,+0.264]$ (10/10) \\
$\varepsilon_{\mathrm{off}}=0.3$ $-$ global $\varepsilon=0.3$ & $-0.039$ $[-0.048,-0.029]$ (0/10) & $-0.031$ $[-0.046,-0.018]$ (1/10) \\
Global $\varepsilon=0.3$ $-$ graph + random fallback & $+0.082$ $[+0.036,+0.128]$ (9/10) & $+0.078$ $[+0.036,+0.123]$ (8/10) \\
$\varepsilon_{\mathrm{off}}=0.3$ $-$ graph + random fallback & $+0.044$ $[+0.001,+0.090]$ (7/10) & $+0.046$ $[+0.006,+0.092]$ (8/10) \\
Hard distilled ($\varepsilon=0$) $-$ graph + random fallback & $-0.211$ $[-0.264,-0.154]$ (0/10) & $-0.241$ $[-0.290,-0.192]$ (0/10) \\
Soft $-$ hard distilled ($\varepsilon=0$) & $-0.283$ $[-0.338,-0.243]$ (0/10) & $-0.253$ $[-0.281,-0.222]$ (0/10) \\
\bottomrule
\end{tabular}}
\end{table}

\begin{table}[t]
\caption{Policy coverage on DoorKey $8\times8$ (150k steps, StructRL with goal prior and graph-action transfer $n=4$ and DQN baseline $n=4$, seeds 2100--2103). $C_\theta$ is evaluated on a fixed random sample of 600 of the reachable states of each layout, identical for all arms of a seed; mean $\pm$ standard deviation over seeds.}
\label{tab:cov8}
\centering\footnotesize
\begin{tabular}{@{}lc@{}}
\toprule
Controller & $C_\theta$ \\
\midrule
Plain DQN (online) & 0.00 $\pm$ 0.00 \\
Online Q-model & 0.22 $\pm$ 0.10 \\
Hard distilled, $\varepsilon=0$ & 0.41 $\pm$ 0.06 \\
Hard distilled, $\varepsilon=0.1$ & 0.75 $\pm$ 0.04 \\
Hard distilled, $\varepsilon=0.3$ & 0.86 $\pm$ 0.04 \\
Hard distilled, $\varepsilon_{\mathrm{off}}=0.3$ & 0.75 $\pm$ 0.04 \\
Soft distilled, $\varepsilon=0$ & 0.14 $\pm$ 0.02 \\
Graph + random fallback & 0.72 $\pm$ 0.08 \\
\bottomrule
\end{tabular}
\end{table}

\begin{table}[t]
\caption{Randomization control on DoorKey $6\times6$ (60k steps, seeds 2000--2009, ordinary StructRL: 2000--2004; rerun of the training; $C_\theta$ as in Table~\ref{tab:cov}, mean $\pm$ standard deviation over seeds). Random labels: the student is distilled on one randomly chosen supported action per graph node instead of the dynamic-programming action. Plain DQN is shown with the same $\varepsilon$.}\label{tab:eps}
\centering\small\begin{tabular}{lccc}\toprule
Controller & StructRL + goal prior + graph transfer & StructRL & DQN baseline \\\midrule
Graph + random fallback & 0.84 $\pm$ 0.05 & 0.83 $\pm$ 0.02 & -- \\
Online Q-model, $\varepsilon=0$ & 0.45 $\pm$ 0.10 & 0.11 $\pm$ 0.02 & 0.04 $\pm$ 0.05 \\
Online Q-model, $\varepsilon=0.1$ & 0.72 $\pm$ 0.10 & 0.41 $\pm$ 0.09 & 0.23 $\pm$ 0.19 \\
Online Q-model, $\varepsilon=0.3$ & 0.84 $\pm$ 0.09 & 0.60 $\pm$ 0.11 & 0.38 $\pm$ 0.27 \\
Online Q-model, $\varepsilon_{\mathrm{off}}=0.3$ & 0.69 $\pm$ 0.10 & 0.28 $\pm$ 0.07 & -- \\
Distilled, random labels, $\varepsilon=0$ & 0.02 $\pm$ 0.01 & 0.02 $\pm$ 0.01 & -- \\
Distilled, random labels, $\varepsilon=0.3$ & 0.24 $\pm$ 0.11 & 0.27 $\pm$ 0.09 & -- \\
Hard distilled, $\varepsilon=0$ & 0.59 $\pm$ 0.09 & 0.62 $\pm$ 0.02 & -- \\
Hard distilled, $\varepsilon=0.3$ & 0.91 $\pm$ 0.07 & 0.90 $\pm$ 0.05 & -- \\
Uniform random actions & 0.41 $\pm$ 0.14 & 0.38 $\pm$ 0.12 & 0.41 $\pm$ 0.14 \\
\bottomrule\end{tabular}
\end{table}

\section{Proofs of the structural guarantees}
\label{app:proofs}
The proofs removed from the main text are given here for completeness.

\paragraph{Proof of Proposition~\ref{prop:abstract}.}
Conditions (i) and (iii) give an exact MDP abstraction for the reward objective and make $\bar P$ and $\bar r$ well defined; condition (ii) ensures that the anchor criterion is representable at the abstract level. Let $V_0=0$ and $\bar V_0=0$, so $V_0(s)=\bar V_0(\phi(s))$. If $V_k(s')=\bar V_k(\phi(s'))$ for all $s'$, then for every $s$ with $\phi(s)=z$ and every action $a$, $r(s,a)+\gamma\sum_{s'}P(s'\mid s,a)V_k(s')=\bar r(z,a)+\gamma\sum_{z'}\bar P(z'\mid z,a)\bar V_k(z')$, because the sum over $s'$ can be grouped by $\phi(s')=z'$; hence $V_{k+1}(s)=\bar V_{k+1}(\phi(s))$. Value iteration converges for $\gamma<1$ and for proper finite episodic tasks, so $V^*(s)=\bar V^*(\phi(s))$. The same identity holds for the action values, $Q^*(s,a)=\bar Q^*(\phi(s),a)$; an action that is greedy for $\bar Q^*(\phi(s),\cdot)$ is therefore greedy for $Q^*(s,\cdot)$, and $\pi(s)=\bar\pi(\phi(s))$ is optimal in the original MDP. The anchor is a union of structural states by condition (ii), so the same argument applies to the anchor-relative shortest-path objective with an absorbing anchor.

\paragraph{Proof of Proposition~\ref{prop:support}.}
Every edge of $\G$ represents an observed transition. Under deterministic successor consistency and anchor consistency, applying its recorded action at any state with the same structural key reaches the recorded structural successor. Induction over the length of a finite observed path gives structural executability up to the absorbing anchor: for a path of length one the claim is the consistency statement itself, and if the first $n$ actions reach the $n$-th structural state of the path from any represented state, the $(n+1)$-st recorded action reaches the next structural state by the same statement. Executability is structural: the recorded actions lead from every represented state through the recorded sequence of structural states to the anchor, but the original sequence of environment states need not be repeated. The set of observed paths is a subset of feasible environment paths, so the minimum length over observed paths is no smaller than the minimum over all feasible environment paths. Unit-cost dynamic programming on the finite observed graph finds the former minimum. This does not prove that reward-maximizing Eq.~\eqref{eq:graph-value} chooses that path. The claim also requires the environment state used in the proposition to contain any variables needed for deterministic execution; omitted time limits or hidden variables can invalidate its premise.

\section{Boundary and off-support diagnostics}
\label{app:boundary}
On PointMaze Large the transition consistency $C_\phi$ is 0.59 for position cells of side 0.5 and 0.64--0.69 for the alternative tested projections; only 13--36\% of counted state-action pairs have a single observed successor. On DoorKey the corresponding quantity is 1.0. In the DoorKey perturbed-start control, training through the random prefix raises the share of far starts with a graph path from 0.88 to 0.93, but the change in the off-support Q-model disadvantage is not distinguishable from zero. These results are treated as boundary diagnostics rather than evidence of generalization beyond graph support.

\section{Further variants}
\label{app:variants}
Several further variants were tried and are not part of the method. Inserting graph values into DQN targets (graph bootstrap), an anchor-local residual value head and shaping-based transfer gave no stable gain over StructRL; the positive transfer result of the paper therefore concerns action preferences, not calibrated values. Extra episodes that navigate over observed edges to frontier states and explore randomly there raised the coverage of the hard-distilled Q-model from 0.53 up to 0.93 in a three-seed run (\texttt{struct\_mgkahg}, DoorKey $6\times6$), but no variant met the adoption rule fixed for it (coverage $\ge0.85$ in all seeds, graph at most 70\% of the reachable states, at least 0.05 above random targets); a separate recovery policy did not meet its rule either. Further small variants (L2 anchoring of the margin, softmax and path-replay margin variants, moderate action and observation noise, a negative-log-likelihood actor loss for SAC) gave mixed small-sample results and none beyond the main configuration.

\section{Run protocol and decision rules}
\label{app:protocol}
Each experiment block was specified before its runs in a text file giving task, seeds, arms, budget, measures and decision rule; the SHA-256 checksum of each file was stored at the time of writing, and the files are kept with the code. Runs were executed in time-limited queue windows and resumed after each window; the windows concern only the execution and not the environment-step budgets. Results are counts per task and are not pooled across tasks. The numerical tables of Section~\ref{sec:results} and the appendix can be regenerated from the stored result files, and the tables of coverage, randomization and training budget (Table~\ref{tab:long}) are regenerated and compared automatically; the full training runs of all tables were not repeated by a second party. Stable success is the first of two consecutive evaluation checkpoints with success 1; there are 25 checkpoints per run with one deterministic episode each. Differences are paired by seed with bootstrap 95\% intervals (5000 draws). Table~\ref{tab:protocol} lists the blocks whose results are reported in this paper.

\begin{table}[!t]
\caption{Experiment blocks and decision rules (summary).}\label{tab:protocol}
\centering\footnotesize
\begin{tabularx}{\linewidth}{p{3.1cm}p{2.6cm}X}
\toprule
Block & Seeds & Decision rule (summary) \\
\midrule
\multicolumn{3}{l}{\emph{Nuisance and transfer study}}\\
Two-branch confirmation & 1710--1719 & Two-branch stable success in $\ge7/10$ seeds, joint Q-model $\le2/10$, paired AUC interval excludes 0. \\
Task size, nuisance, DoorKey $8\times8$ & 1720--1724; 1730--1734 & Descriptive; nuisance dimensions 0, 100, 5000. \\
DP head; gradient cosine and detach & 1740--1744; 1760--1764 & Adopt only if clearly better than the plain two-branch Q-model; gradient conflict only if the joint Q-model shows more negative cosines than the two-branch Q-model and its permuted control. \\
Graph necessity, $6\times6$ / $8\times8$ & 1780--1784; 1790--1799 & Graph episodes necessary if plain is stable in at most 3 of 10 seeds and \texttt{struct} in at least 8 of 10. \\
Automatic selection, versions 1--3 & 1800--1804; 1810--1814; 1820--1824 & $\ge4/5$ stable and median AUC within 0.1 of the hand-built $\phi$. \\
\multicolumn{3}{l}{\emph{TER and graph versus network}}\\
TER comparison & 1100--1109; 1200--1204; 1300--1304 & D1--D6 paired by seed; lower bound $>0$ better, upper bound $<0$ worse, else no detectable difference. \\
DQN hybrid; random start (A, B, C); prefix exploration (B2) & 1400--1409 & $\Delta_{\mathrm{off}}$ over seeds with $\ge5$ off-support starts; graph-path share and Q-model success on far starts; same reading rule. \\
SAC random start & 700--709; 800--801 (larger map) & Added value off support if $\Delta_{\mathrm{off}}\ge0.2$ in $\ge3$ anchored seeds and mean $\ge0.1$; main seeds of the larger map only if one of the first two seeds reaches the goal. \\
\multicolumn{3}{l}{\emph{Policy-coverage study}}\\
Policy-coverage study & 2000--2009 ($6\times6$); 2100--2103 ($8\times8$) & Distillation helps if the hard-distilled Q-model exceeds the online Q-model in $C_\theta$ in at least 9 of 10 seeds per arm; randomization helps if $\varepsilon=0.3$ exceeds $\varepsilon=0$ in at least 9 of 10 seeds; the support-aware controller replaces the global one if its mean $C_\theta$ is at most 0.03 lower and it needs at most 1.05 times the steps on the common set; soft targets descriptive; the scaling statement for $8\times8$ holds if the final controller exceeds plain DQN by at least 0.2 in all four seeds and the online Q-model in at least three. \\
Randomization control & 2000--2009 (\texttt{struct\_mgkahg}, plain); 2000--2004 (\texttt{struct}) & E1: distillation adds at least $+0.10$ in $C_\theta$ at equal $\varepsilon=0.3$ in at least 9 of 10 seeds. E2: graph labels exceed random labels by at least $+0.10$ in at least 9 of 10 seeds. \texttt{struct} descriptive. \\
\bottomrule
\end{tabularx}
\end{table}

\paragraph{Deviations and run order.} (i) The first four runs of the DP-head block were run before the transfer-residual logging was added and were repeated; the earlier runs are kept separately. (ii) One run was terminated by the operating system for lack of memory and repeated with the same seed. (iii) The specification of the automatic selection was amended twice after diagnosed failures of versions 1 and 2 (the amendments are marked in the file and its checksum was renewed); the final rule is therefore a test only on its new seeds. (iv) The B2 seeds 1401--1404 were started in parallel before the preceding random-start block had finished, with the specified code and without inspection of any result; they and arm A seeds 1400--1406 were rerun with logging of the start-support flags, and the reported values come from these reruns. Earlier outputs of the superseded code are kept separately.

\paragraph{Randomization control.} The control reruns the training of the policy-coverage seeds with additional controllers (Section~\ref{sec:coverage}); the shared rows agree with Table~\ref{tab:cov} within 0.01. The specification file was amended once, after the first ten runs had started and before any result was evaluated, to add the arm \texttt{struct} with five seeds; E1 and E2 apply to \texttt{struct\_mgkahg} with ten seeds, and \texttt{struct} is descriptive.

\paragraph{Policy-coverage study.} The 38 runs of the study were executed in queue windows of 90 minutes and resumed after each window; no run was excluded, and every run of the specified list is reported. The number of sampled states for DoorKey $8\times8$ (600) was fixed in the specification file before the first run. The steps in Table~\ref{tab:cov} are computed on the common set of all six controllers; the step criterion compares two controllers on their common set and gives the same decision.

\section{Implementation details of the two-branch Q-model}
\label{app:phi2}
\paragraph{Feature-access controls.} Besides the two-branch model, the controls include a structural-only Q-model, a joint-input model with $\phi(s)$ appended, a detached-raw-branch variant in which the graph/margin step updates only the structural branch, and an explicit DP-head variant with optional confidence weighting. These controls are not part of the main StructRL procedure; they isolate why graph-derived labels fail under high-dimensional nuisance input. For one control we also log the cosine between the gradients of the TD loss and of the graph/margin loss per parameter block, with a permuted-label control.

Both branches are single linear layers with ReLU of width 128; their concatenation (256 units) feeds the dueling value/advantage trunk of the base DQN. The structural branch receives the observation coordinates that constitute $\phi(s)$ (hand-built) or the coordinates kept by the automatic selection; the raw branch receives the complete observation including all nuisance dimensions. All other hyperparameters equal Table~\ref{tab:hyper-dqn}. The automatic selection uses 5000 additional random-action steps of the same task and layout, which are not counted in the training budget; the bias-corrected between-key share $R^2_j$ is computed in double precision over keys visited at least twice.

\paragraph{Input selection and refinement.} Among many observation dimensions, a dimension is discarded only with positive evidence of nuisance: after 5000 random-action steps, its within-key variance over the structural keys visited at least twice is at least $10^{-4}$ and the bias-corrected between-key share $R_j^2$ of its variance is below 0.5. The remaining dimensions form the input of the structural branch. In a controlled setting with 1000 synthetic nuisance dimensions the final rule (version 3) reaches stable success in 5 of 5 seeds with median AUC 0.72, against 0.92 for the hand-built $\phi$ on the same seeds. Version 1 reached stable success in 0 of 5 seeds because it dropped dimensions that were constant during the random warm-up; version 2 succeeded in 4 of 5 seeds and failed on a dimension that varied only in singly visited keys. The rule was corrected after each failure, so the final rule was run on new seeds only. It needs the structural key and forms no new features. For refinement after a detected inconsistency, stored raw transitions are re-aggregated under a finer projection from a predefined ladder; in a three-seed online experiment this reached AUCs of 0.24, 0.52 and 0.32 (mean 0.36; hand-built exact $\phi$ 0.49, fixed coarse $\phi$ 0), switching at the earliest checks and without reducing the graph size. It shows that re-aggregation is possible, not that adaptive abstraction improves efficiency.

\section{Hyperparameters}
\begin{table}[!htbp]
\caption{DQN base-learner and StructRL hyperparameters. Shared settings apply to all DQN arms; prior and margin parameters apply only to the indicated arms.}\label{tab:hyper-dqn}
\centering\footnotesize
\begin{tabular}{ll}
\toprule
Parameter & Value \\
\midrule
\multicolumn{2}{l}{\emph{DQN base learner: shared}}\\
Observation & fully observed grid, scaled by $1/10$, flattened\\
Network & 2 hidden layers of 256 units, dueling heads, ReLU\\
Update rule & Double DQN, $n$-step return with $n=10$\\
Optimizer & Adam, learning rate $5\cdot10^{-4}$\\
Minibatch size & 64\\
Replay memory size & $T$ (all transitions)\\
Learning starts & 1\,000 steps\\
Training frequency & one gradient step every 4 environment steps\\
Target network update interval & 500 steps\\
Discount $\gamma$ & 0.99\\
Exploration $\epsilon$ & 1.0 $\to$ 0.05, linear over first $0.3\,T$\\
\midrule
\multicolumn{2}{l}{\emph{StructRL: shared graph settings}}\\
Graph episode probability $p_{\mathrm{graph}}$ & 0.5\\
Guidance length $H_0$ & 100 steps, linear decay to 0 at $0.6\,T$\\
Random action probability in guidance $\epsilon_G$ & 0.1\\
Support threshold $n_{\min}$ & 1\\
Graph discount $\gamma$ & 0.99\\
Frontier novelty scale $\beta$, discount $\gamma_\nu$ & 1.0, 0.95\\
\midrule
\multicolumn{2}{l}{\emph{Known-anchor prior arms only}}\\
Prior multiplier & $w_A(z)=1+3\exp(-d_A(z)/3)$\\
Prior strength, scale & $3.0,\ \sigma=3.0$\\
Applies to & \texttt{struct\_ka}, \texttt{struct\_mgkahg}, \texttt{struct\_kawrong}\\
\midrule
\multicolumn{2}{l}{\emph{Margin arms only}}\\
Margin $m$, weight $\lambda$ & $0.8,\ 1.0$\\
Applies to & \texttt{struct\_mghg}, \texttt{struct\_mgshufhg}, \texttt{struct\_mgkahg}\\
\midrule
\multicolumn{2}{l}{\emph{Evaluation}}\\
Checkpoints & 25, every $T/25$ steps\\
Evaluation episodes per checkpoint & 1 (deterministic policy)\\
Layout & fixed per seed (layout $=$ seed)\\
\bottomrule
\end{tabular}
\end{table}

\begin{table}[!htbp]
\caption{Core SAC base-learner and StructRL hyperparameters for PointMaze M and Large. The known-anchor prior is arm-specific and is analyzed separately.}\label{tab:hyper-sac}
\centering\footnotesize
\resizebox{\linewidth}{!}{\begin{tabular}{ll}
\toprule
Parameter & Value \\
\midrule
\multicolumn{2}{l}{\emph{SAC base learner}}\\
Observation & position, velocity, goal position (6-dim.)\\
Networks & actor and two critics, 2 hidden layers of 256 units, ReLU\\
Optimizer & Adam, learning rate $3\cdot10^{-4}$ (actor, critics, entropy coefficient)\\
Minibatch size & 256\\
Replay memory size & $T$ (all transitions)\\
Learning starts & 1\,000 steps\\
Training frequency & one gradient step per environment step\\
Target smoothing $\tau$ & 0.005, every step\\
Discount $\gamma$ & 0.99\\
Entropy coefficient $\alpha$ & automatic, initial value 1.0, target entropy $-\dim\mathcal A=-2$\\
Episode length & 600 steps (M), 1\,000 steps (Large)\\
Training budget $T$ & 80k steps, five seeds per arm\\
\midrule
\multicolumn{2}{l}{\emph{StructRL}}\\
Structural state $\phi$ & grid cell of side 0.5 over the position\\
Action prototypes $K$ & 9 (zero action and 8 compass directions, $\|u\|_\infty=1$)\\
Prototype noise & Gaussian, std.\ $0.1\,(a_{\max}-a_{\min})$, clipped to the action box\\
Block length $H$ & 10 steps\\
Frontier probability $p_{\mathrm{explore}}$ & 0.5 per block (1.0 before learning starts)\\
Frontier: novelty scale $\beta$, discount $\gamma_\nu$ & 1.0, 0.95\\
Frontier replanning & every 500 steps and at episode end\\
Support threshold $n_{\min}$ (planning graph) & 2\\
Anchor & first transition with $r>0.5$\\
Graph-direction loss weight $\lambda$ & 1.0\\
\midrule
\multicolumn{2}{l}{\emph{Evaluation}}\\
Checkpoints & every 10k steps (8 per run)\\
Evaluation episodes per checkpoint & 10 (deterministic policy, fixed reset seeds)\\
Held-out map & 20 checkpoints, 5 episodes each, 20 episodes at the end\\
Success & goal reached within the episode\\
\bottomrule
\end{tabular}}
\end{table}

\section{Implementation details}
\label{app:impl}
\paragraph{Graph and support.} The graph stores for every structural state $z$, action representation $u$ and successor $z'$ the count $N(z,u,z')$, the reward sum and the number of terminal transitions. A pair $(z,u)$ is \emph{supported} if $N(z,u)\ge n_{\min}$, with $n_{\min}=1$ for DQN (deterministic structural keys), and $n_{\min}=2$ for the SAC frontier graph on PointMaze. Unsupported pairs have no value and no label. Terminal transitions bootstrap with zero; time-limit truncations are not terminal. Successor nodes without a supported action have value zero.

\paragraph{Graph policy.} $\hat Q$ is computed by value iteration over the supported pairs until the largest change is below $10^{-7}$ (at most 5000 sweeps). $\pi_G(z)$ is the arg\,max over the supported actions of $z$ for $z\in\mathcal Z_A$ (first index in case of ties). $\mathcal Z_A$ is found by a backward search from the anchor transitions. Margin loss, SAC direction loss and hard distillation use labels on $\mathcal Z_A$ only. The executed controller $\pi_{\mathrm{exec}}$ takes the arg\,max at every supported state; outside $\mathcal Z_A$ the values are zero and it takes the first supported action. If $z$ has no supported action, $\pi_G$ is undefined and the learner acts (DQN) or the sample is not part of $\mathcal B_G$ (SAC).

\paragraph{Frontier policy.} $Q_\nu$ is initialized with $\beta$ at every node (the value of an untried action) and iterated for at most 200 sweeps with tolerance $10^{-4}$; it is recomputed from the current counts at every planning step. The frontier action is the arg\,max of $Q_\nu(z,\cdot)$ with random tie-breaking (noise of order $10^{-9}$). If the current state is not yet in the graph, the action is drawn uniformly.

\paragraph{DQN guidance and goal prior.} In graph-guided DQN episodes, guidance follows the graph policy for $h(t)=H_0\max(0,1-t/(0.6T))$ steps after a supported anchor path exists; guided actions are replaced by a uniformly random action with probability $\epsilon_G$. In the known-goal experiments the frontier novelty signal is multiplied before first anchor contact by $w_A(z)=1+3\exp(-d_A(z)/3)$, where $d_A$ is Manhattan distance to the goal. The DQN margin uses $m=0.8$ and $\lambda=1.0$; the shuffled-label control permutes graph-optimal action sets within equal anchor-distance shells.

\paragraph{Planning schedule.} DQN: once at the start of every episode. SAC: every 500 environment steps and at the end of every episode. The graph is updated after every environment step; the structural key of a transition is $(\phi(s_t),\psi(a_t),\phi(s_{t+1}))$ in DQN, whereas SAC records a transition only if the structural cell changes or the episode ends.

\paragraph{Structural states and observations.} DQN: $\phi(s)$ is the tuple (agent position, agent direction, carried object, checksum of the encoded grid); the Q-model input is the fully observed grid encoding divided by 10 and flattened. SAC: $\phi(s)$ is the cell of side $0.5$ over the two position coordinates; the Q-model input is position, velocity and goal position (six values).

\section{Environment details}
\label{app:envs}
\paragraph{MiniGrid.} We use \texttt{FullyObsWrapper} and \texttt{ImgObsWrapper}; the observation is the full grid with three channels (object type, color, state) including the agent. There are seven discrete actions (turn left, turn right, move forward, pick up, drop, toggle, done). The reward is zero until the task is completed and then $1-0.9\,t/t_{\max}$, where $t_{\max}$ is 360 (DoorKey 6x6), 270 (KeyCorridor S3R2, S3R3) or 288 (ObstructedMaze 1Dl, UnlockPickup). An episode ends on success or after $t_{\max}$ steps. The layout is drawn from the seed and is the same in every episode of a run. In the original DQN performance block the anchor is registered at the first positive reward during training.

\paragraph{PointMaze.} We use \texttt{PointMaze\_UMaze-v3} (gymnasium-robotics) with custom maze maps. The agent is a ball with a two-dimensional action in $[-1,1]^2$ (force); the observation contains position, velocity and goal position. The reward is 1 on reaching the goal cell and the episode then ends. Episode length is 600 steps (M) and 1000 steps (Large); the goal is not resampled between episodes. In the reported SAC experiments the anchor is the first goal contact.

\paragraph{SearchGrid.} An $n\times n$ grid with random walls, a start cell and $m$ ordered subgoal cells; five actions (four moves and stay). A subgoal is registered when the agent is on its cell (with probability $d$ per step; $d=1$ exact). The last registration gives reward 1. The observation contains position, the number of registered subgoals and, optionally, nuisance bits that are redrawn at every step. Episodes are truncated after a fixed horizon; the budget is the total number of environment steps.


\begin{thebibliography}{99}
\bibitem{rusu2016}
A. A. Rusu, S. G. Colmenarejo, C. Gulcehre, G. Desjardins, J. Kirkpatrick, R. Pascanu, V. Mnih, K. Kavukcuoglu, and R. Hadsell.
Policy distillation.
In \emph{International Conference on Learning Representations (ICLR)}, 2016. arXiv:1511.06295.

\bibitem{hinton2015}
G. Hinton, O. Vinyals, and J. Dean.
Distilling the knowledge in a neural network.
arXiv:1503.02531, 2015.

\bibitem{givan2003}
R. Givan, T. Dean, and M. Greig.
Equivalence notions and model minimization in Markov decision processes.
\emph{Artificial Intelligence}, 147(1--2):163--223, 2003.

\bibitem{ravindran2003}
B. Ravindran and A. G. Barto.
SMDP homomorphisms: an algebraic approach to abstraction in semi-Markov decision processes.
In \emph{Proceedings of the International Joint Conference on Artificial Intelligence (IJCAI)}, 2003.

\bibitem{li2006}
L. Li, T. J. Walsh, and M. L. Littman.
Towards a unified theory of state abstraction for MDPs.
In \emph{International Symposium on Artificial Intelligence and Mathematics (ISAIM)}, 2006.

\bibitem{yamauchi1997}
B. Yamauchi.
A frontier-based approach for autonomous exploration.
In \emph{Proceedings of the IEEE International Symposium on Computational Intelligence in Robotics and Automation (CIRA)}, pages 146--151, 1997.

\bibitem{bertsekas1996}
D. P. Bertsekas.
\emph{Dynamic Programming and Optimal Control}.
Athena Scientific, 1996.

\bibitem{lin1992}
L.-J. Lin.
Self-improving reactive agents based on reinforcement learning, planning and teaching.
\emph{Machine Learning}, 8:293--321, 1992.

\bibitem{schaul2016}
T. Schaul, J. Quan, I. Antonoglou, and D. Silver.
Prioritized Experience Replay.
In \emph{International Conference on Learning Representations}, 2016.

\bibitem{hong2022ter}
Z.-W. Hong, T. Chen, Y.-C. Lin, J. Pajarinen, and P. Agrawal.
Topological Experience Replay.
In \emph{International Conference on Learning Representations}, 2022.

\bibitem{moore1993}
A. W. Moore and C. G. Atkeson.
Prioritized sweeping: Reinforcement learning with less data and less time.
\emph{Machine Learning}, 13:103--130, 1993.

\bibitem{sutton1991}
R. S. Sutton.
Dyna, an integrated architecture for learning, planning, and reacting.
\emph{SIGART Bulletin}, 2(4):160--163, 1991.

\bibitem{eysenbach2019}
B. Eysenbach, R. Salakhutdinov, and S. Levine.
Search on the Replay Buffer: Bridging Planning and Reinforcement Learning.
In \emph{International Conference on Learning Representations}, 2019.

\bibitem{savinov2018}
N. Savinov, A. Dosovitskiy, and V. Koltun.
Semi-Parametric Topological Memory for Navigation.
In \emph{International Conference on Learning Representations}, 2018.

\bibitem{emmons2020}
S. Emmons, A. Jain, M. Laskin, T. Kurutach, P. Abbeel, and D. Pathak.
Sparse Graphical Memory for Robust Planning.
In \emph{Advances in Neural Information Processing Systems}, 2020.

\bibitem{zhu2020erlam}
G. Zhu, Z. Lin, G. Yang, and C. Zhang.
Episodic Reinforcement Learning with Associative Memory.
In \emph{International Conference on Learning Representations}, 2020.

\bibitem{hoppe2020qgraph}
S. Hoppe and M. Toussaint.
Qgraph-bounded Q-learning: Stabilizing Model-Free Off-Policy Deep Reinforcement Learning.
arXiv:2007.07582, 2020.

\bibitem{hoppe2020industrial}
S. Hoppe, M. Giftthaler, R. Krug, and M. Toussaint.
Sample-Efficient Learning for Industrial Assembly using Qgraph-bounded DDPG.
In \emph{IEEE/RSJ International Conference on Intelligent Robots and Systems}, 2020.

\bibitem{haarnoja2018}
T. Haarnoja, A. Zhou, P. Abbeel, and S. Levine.
Soft Actor-Critic: Off-Policy Maximum Entropy Deep Reinforcement Learning with
a Stochastic Actor.
In \emph{Proceedings of the 35th International Conference on Machine Learning}, 2018.

\bibitem{hester2018dqfd}
T. Hester, M. Vecerik, O. Pietquin, M. Lanctot, T. Schaul, B. Piot, D. Horgan, J. Quan, A. Sendonaris, G. Dulac-Arnold, I. Osband, J. Agapiou, J. Z. Leibo, and A. Gruslys.
Deep Q-learning from Demonstrations.
In \emph{Proceedings of the AAAI Conference on Artificial Intelligence}, 2018. arXiv:1704.03732.

\bibitem{vecerik2017ddpgfd}
M. Vecerik, T. Hester, J. Scholz, F. Wang, O. Pietquin, B. Piot, N. Heess, T. Roth{\"o}rl, T. Lampe, and M. Riedmiller.
Leveraging Demonstrations for Deep Reinforcement Learning on Robotics Problems with Sparse Rewards.
arXiv:1707.08817, 2017.

\bibitem{nair2020awac}
A. Nair, A. Gupta, M. Dalal, and S. Levine.
AWAC: Accelerating Online Reinforcement Learning with Offline Datasets.
arXiv:2006.09359, 2020.

\bibitem{ecoffet2021goexplore}
A. Ecoffet, J. Huizinga, J. Lehman, K. O. Stanley, and J. Clune.
First return, then explore.
\emph{Nature}, 590:580--586, 2021.

\bibitem{blundell2016mfec}
C. Blundell, B. Uria, A. Pritzel, Y. Li, A. Ruderman, J. Z. Leibo, J. Rae, D. Wierstra, and D. Hassabis.
Model-Free Episodic Control.
arXiv:1606.04460, 2016.

\bibitem{lee2019ebu}
S. Y. Lee, S. Choi, and S.-Y. Chung.
Sample-Efficient Deep Reinforcement Learning via Episodic Backward Update.
In \emph{Advances in Neural Information Processing Systems}, 2019.

\bibitem{bellemare2016pseudocount}
M. G. Bellemare, S. Srinivasan, G. Ostrovski, T. Schaul, D. Saxton, and R. Munos.
Unifying Count-Based Exploration and Intrinsic Motivation.
In \emph{Advances in Neural Information Processing Systems}, 2016.


\bibitem{mnih2015dqn}
V. Mnih, K. Kavukcuoglu, D. Silver, A. A. Rusu, J. Veness, M. G. Bellemare,
A. Graves, M. Riedmiller, A. K. Fidjeland, G. Ostrovski, S. Petersen,
C. Beattie, A. Sadik, I. Antonoglou, H. King, D. Kumaran, D. Wierstra,
S. Legg, and D. Hassabis.
Human-level control through deep reinforcement learning.
\emph{Nature}, 518:529--533, 2015.

\bibitem{chevalierboisvert2023minigrid}
M. Chevalier-Boisvert, B. Dai, M. Towers, R. Perez-Vicente, L. Willems, S. Lahlou, S. Pal, P. S. Castro, and J. Terry.
Minigrid \& Miniworld: Modular \& Customizable Reinforcement Learning Environments for Goal-Oriented Tasks.
In \emph{Advances in Neural Information Processing Systems, Datasets and Benchmarks Track}, 2023.

\bibitem{gymnasiumrobotics2023}
R. de Lazcano, K. Andreas, J. J. Tai, S. R. Lee, and J. Terry.
Gymnasium-Robotics.
Farama Foundation, 2023. \url{https://github.com/Farama-Foundation/Gymnasium-Robotics}.


\bibitem{plappert2018multigoal}
M. Plappert, M. Andrychowicz, A. Ray, B. McGrew, B. Baker, G. Powell, J. Schneider, J. Tobin, M. Chociej, P. Welinder, V. Kumar, and W. Zaremba.
Multi-Goal Reinforcement Learning: Challenging Robotics Environments and Request for Research.
arXiv:1802.09464, 2018.

\bibitem{andrychowicz2020dexterous}
M. Andrychowicz, B. Baker, M. Chociej, R. J\'ozefowicz, B. McGrew, J. Pachocki, A. Petron, M. Plappert, G. Powell, A. Ray, J. Schneider, S. Sidor, J. Tobin, P. Welinder, L. Weng, and W. Zaremba.
Learning dexterous in-hand manipulation.
\emph{The International Journal of Robotics Research}, 39(1):3--20, 2020.

\bibitem{tang2025robotics}
C. Tang, B. Abbatematteo, J. Hu, R. Chandra, R. Mart\'in-Mart\'in, and P. Stone.
Deep Reinforcement Learning for Robotics: A Survey of Real-World Successes.
\emph{Annual Review of Control, Robotics, and Autonomous Systems}, 8:153--188, 2025.

\bibitem{mayerhoff2026ppc}
J. Mayerhoff and M. Schmidt.
Reinforcement learning for autonomous production planning and control: A systematic literature review.
\emph{Journal of Manufacturing Systems}, 86:546--568, 2026.

\bibitem{florensa2017reverse}
C. Florensa, D. Held, M. Wulfmeier, M. Zhang, and P. Abbeel.
Reverse Curriculum Generation for Reinforcement Learning.
In \emph{Proceedings of the 1st Annual Conference on Robot Learning}, PMLR 78:482--495, 2017.

\bibitem{resnick2018backplay}
C. Resnick, R. Raileanu, S. Kapoor, A. Peysakhovich, K. Cho, and J. Bruna.
Backplay: ``Man muss immer umkehren''.
arXiv:1807.06919, 2018.

\end{thebibliography}
\end{document}